\documentclass[10pt,final,journal]{IEEEtran} 

\usepackage{amsmath,amsfonts}
\usepackage{algorithm}
\usepackage{array}
\usepackage[caption=false,font=normalsize,labelfont=sf,textfont=sf]{subfig}
\usepackage{textcomp}
\usepackage{stfloats}
\usepackage{url}
\usepackage{verbatim}
\usepackage{graphicx}
\usepackage{cite}
\usepackage{amsmath}
\usepackage{multirow}
\usepackage{xcolor}
\usepackage{diagbox}
\usepackage{colortbl}
\usepackage{arydshln}
\usepackage{booktabs}
\usepackage{algpseudocode} 
\usepackage{makecell}

\definecolor{lightred}{RGB}{255, 200, 200}  
\definecolor{lightblue}{RGB}{200, 220, 255} 
\definecolor{mycolor}{rgb}{0.902,0.902,0.980}
\definecolor{darkgreen}{rgb}{0.2,0.8,0.25}

\definecolor{mycolor}{rgb}{0.902,0.902,0.980}
\usepackage{todonotes}

\usepackage{CJKutf8}   

\usepackage{hyperref}
\hypersetup{hypertex=true,
colorlinks=true,
linkcolor=blue,
anchorcolor=blue,
citecolor=blue}

\ifCLASSINFOpdf
\else
\fi

\begin{document}

\title{FusionCounting: Robust visible-infrared image fusion guided by crowd counting via multi-task learning}
\author{
	He Li, Xinyu Liu, Weihang Kong, Xingchen Zhang$^*$,\IEEEmembership{~Member, IEEE} \thanks{This paper was supported partly by the National Natural Science Foundation of China (No. 62306264, 62173290), the Central Government Guided Local Funds for Science and Technology Development of China (No. 236Z0303G), the Natural Science Foundation of Hebei Province in China (No. F2024203091, F2025203045), the Royal Society Research Grant (No. RG\textbackslash{}R1\textbackslash{}251462).

He Li, Xinyu Liu and Weihang Kong are with School of Information Science and Engineering, Yanshan University, Qinhuangdao 066004, China. (Email: lihe@ysu.edu.cn; lxy620@stumail.ysu.edu.cn; whkong@ysu.edu.cn)

Xingchen Zhang is with the Fusion Intelligence Laboratory, Department of Computer Science, University of Exeter, EX4 4RN, United Kingdom. (Email:x.zhang12@exeter.ac.uk)
\newline	$^*$ Corresponding author: Xingchen Zhang 
}
	
}

\markboth{Journal of \LaTeX\ Class Files,~Vol.~XX, No.~XX, 2025}%
{Shell \MakeLowercase{\textit{et al.}}: Bare Demo of IEEEtran.cls for IEEE Journals}

\maketitle

\begin{abstract}
Visible and infrared image fusion (VIF) is an important multimedia task in computer vision. Most visible and infrared image fusion methods focus primarily on optimizing fused image quality.~Recent studies have begun incorporating downstream tasks, such as semantic segmentation and object detection, to provide semantic guidance for VIF. However, semantic segmentation requires extensive annotations, while object detection, despite reducing annotation efforts compared with segmentation, faces challenges in highly crowded scenes due to overlapping bounding boxes and occlusion.~Moreover, although RGB-T crowd counting has gained increasing attention in recent years, no studies have integrated VIF and crowd counting into a unified framework.~To address these challenges, we propose \textbf{FusionCounting}, a novel multi-task learning framework that integrates crowd counting into the VIF process. Crowd counting provides a direct quantitative measure of population density with minimal annotation, making it particularly suitable for dense scenes.~Our framework leverages both input images and population density information in a mutually beneficial multi-task design. To accelerate convergence and balance task contributions, we introduce a dynamic loss function weighting strategy. Furthermore, we incorporate adversarial training to enhance the robustness of both VIF and crowd counting, improving the model’s stability and resilience to adversarial attacks. Experimental results on public datasets demonstrate that FusionCounting not only enhances image fusion quality but also achieves superior crowd counting performance.
\end{abstract}
\begin{IEEEkeywords}
RGB-T crowd counting, image fusion, visible-infrared image fusion, multi-task learning, adversarial training
\end{IEEEkeywords}

\section{Introduction} \label{introduction}
\begin{figure*}
	\centering
	\includegraphics[width=0.98\textwidth]{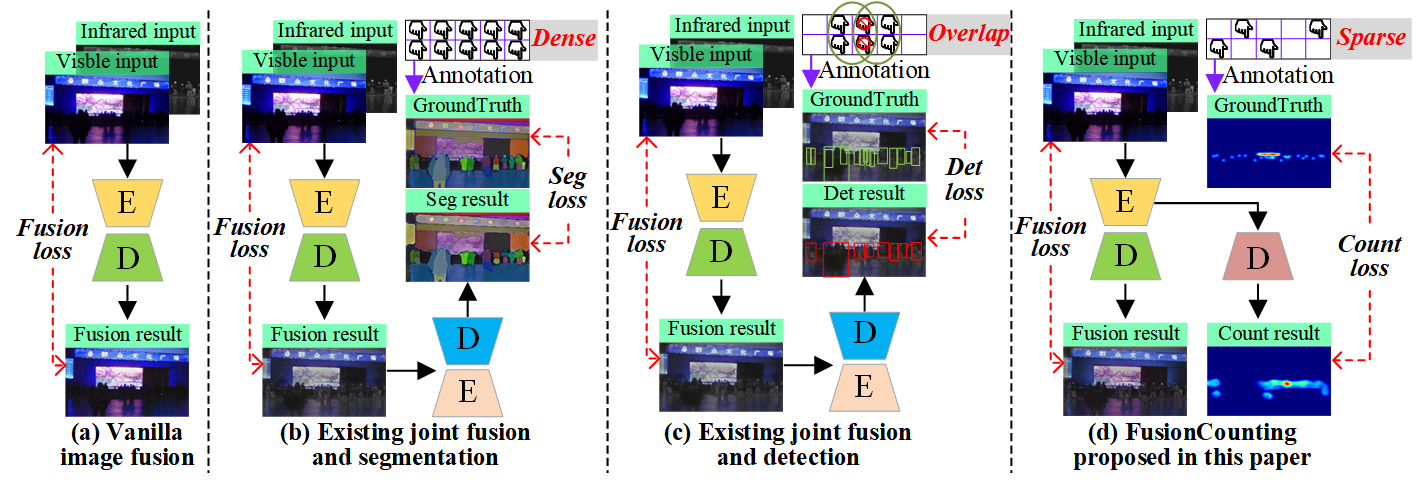}
	\caption{Comparisons of existing methods and the proposed FusionCounting. The rectangles in the top-right corner of (b), (c), and (d) indicate the level of pixel-level annotations required for segmentation, detection, and counting tasks. The bidirectional red dotted lines represent the supervision processes between the estimated results and their labels. `E' and `D' denote encoders and decoders, respectively. Compared to the existing VIF frameworks, FusionCounting enables simultaneous predictions for both tasks with reduced reliance on pixel-level annotations and reduces feature redundancy by using one shared encoder.}
	\label{fig:Figure1}
\end{figure*}
\IEEEPARstart{V}{isible} and infrared image fusion (VIF) \cite{10856398,10814643,zhang2023visible,chen2024hitfusion} refers to the integration of information from visible and infrared images to generate more informative fused images.~Owing to the complementary characteristics of the information from visible images and infrared images, the fused image can be utilized to enhance the performance of various downstream applications, such as object detection, semantic segmentation, and video surveillance.

Because there are no ground-truth labels in the VIF task, most deep learning-based VIF methods are based on unsupervised learning. In these vanilla VIF methods, the loss function (Fusion loss) is obtained by calculating the difference between the generated fused image and its original visible and infrared images, as shown in Fig.~\ref{fig:Figure1}(a). Some studies \cite{tang2022image, zhang2023visible, zhang2024mrfs} have pointed out that optimizing fused image solely is not effective in promoting the performance of downstream vision applications. %
Consequently, some researchers have explored application-oriented VIF methods \cite{tang2022image, liu2022target, zhang2024mrfs}, i.e., consider the downstream vision application in the image fusion process.~A typical example is shown in Fig.~\ref{fig:Figure1}(b), where 
semantic segmentation is utilized to facilitate learning of the image fusion process.
~Some other studies \cite{liu2022target,peng2022mfdetection} utilize object detection as the downstream application, as shown in Fig~\ref{fig:Figure1}(c).  

\begin{figure}[!h]
 \centering
 \includegraphics[width=1\columnwidth]{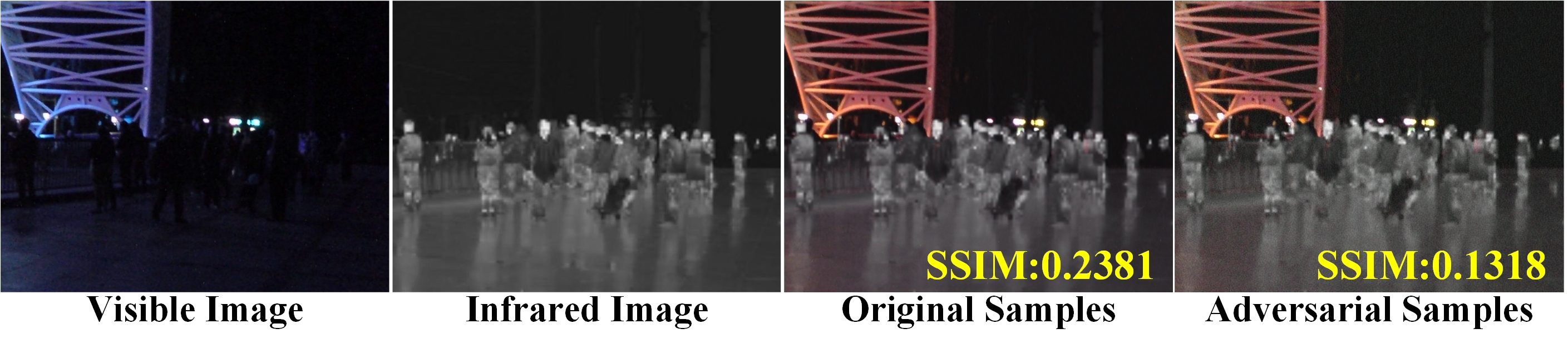}
 \caption{In the VIF task, adversarial attacks severely degrade fusion quality. The fused images are generated using SwinFusion \protect\cite{ma2022swinfusion}. Adversarial examples are generated using the projected gradient descent (PGD) attack \protect\cite{waghela2024robust}.
 }
 \label{fig:attack-training}
\end{figure}

\textbf{However}, semantic segmentation tasks require labor-intensive pixel-level annotations.~Although object detection reduces labeling effort compared to segmentation, it remains prone to annotation errors in highly crowded scenes due to bounding box overlap and occlusion.~\textbf{Moreover}, most existing methods employ a cascaded structure for mutual learning between the two tasks, as shown in Figs.~\ref{fig:Figure1}(b) and (c).~This kind of design can easily lead to redundancy because both the image fusion model and the downstream application model have feature extraction steps, as shown by the two encoders in Figs.~\ref{fig:Figure1} (b) and (c).~\textbf{Furthermore}, existing methods primarily focus on improving task performance. The robustness of these methods, particularly their sensitivity to environmental changes, such as those highlighted by recent focus on adversarial attacks, has been rarely addressed.~Figure \ref{fig:attack-training} illustrates the negative impact of adversarial attacks on the fusion results of the SwinFusion model. As shown, the structural similarity index (SSIM) of the images drops significantly following projected gradient descent (PGD) attacks, highlighting the vulnerability of traditional single-task models to such adversarial perturbations.~\textbf{Finally},  although RGB-T crowd counting has gained increasing attention \cite{liu2021cross,zhang2022spatio,zhou2024mc3,yang2024cagnet}, no studies have integrated VIF and crowd counting into a unified framework to explore the mutual benefits between these two tasks.

To address these issues, this paper integrates crowd counting with VIF into a unified and robust multi-task framework, as shown in Fig.~\ref{fig:Figure1}(d). This framework leverages the synergy between VIF and crowd counting. Although crowd counting is less general than semantic segmentation or object detection, it offers significantly lower supervision costs thanks to sparse point-level annotations and fewer labeling errors. Moreover, supervision from the crowd counting task can guide the fusion process to generate images rich in semantic information, which may benefit various downstream tasks. Specifically, we leverage pedestrian semantic contexts derived from the crowd counting task to guide VIF. To enhance multi-task learning performance, we introduce a dynamic weighting strategy that improves model training. Unlike the Multiple Gradient Descent Algorithm - Upper Bound (MGDA-UB) proposed by Sener and Koltun \cite{sener2018multi}, which computes a Pareto-optimal solution in the tensor space of shared parameters before applying backpropagation, our method dynamically adjusts the weights of multiple task losses and updates all model parameters in an end-to-end manner. Additionally, we design an adversarial training scheme to strengthen the robustness of our model, making it more reliable and precise against malicious attacks.

In summary, the main contributions are as follows:
\begin{itemize}

\item We propose an end-to-end multi-task learning framework for visible-infrared image fusion guided by crowd counting. To our knowledge, our FusionCounting is the first to jointly optimize VIF and crowd counting, making it particularly effective in dense crowd scenarios.

\item We introduce a dynamic loss weighting strategy to balance the contributions of the fusion and counting tasks, accelerating the multi-task learning process.

\item We design an adversarial training scheme to enhance the robustness of FusionCounting against adversarial attacks, ensuring reliable fusion and counting performance. This aspect is often overlooked in prior VIF work.

\item Extensive experiments on two public 
datasets demonstrate the effectiveness of our method in improving image fusion quality, boosting crowd counting performance, and enhancing robustness under adversarial attacks.
\end{itemize}

\section{Related work} \label{related_work}
\subsection{Visible-infrared image fusion}
Visible-infrared image fusion (VIF) \cite{bai2024ibfusion,liu2024infrared,zhao2023cddfuse,zhang2025texture} has been studied for many years as VIF has a wide range of applications, such as object tracking, object objection, semantic segmentation, and salient object detection.~In general, VIF can be divided into conventional methods and deep learning-based methods.~Deep learning-based methods are now dominant in this field, and they can be further classified into CNN-based methods \cite{hu2024pfcfuse}, GAN-based methods \cite{ma2019fusiongan}, transformer-based methods \cite{tang2024itfuse, liu2024stfnet,shi2025frefusion}, and diffusion model-based methods \cite{yue2023dif,yi2024diff}.~Traditionally, performance evaluation in VIF is usually performed in two ways, i.e., quantitative evaluation using image fusion evaluation metrics and qualitative evaluation via human observation.

\begin{figure*}[t]
	\centering
	\includegraphics[width=\textwidth]{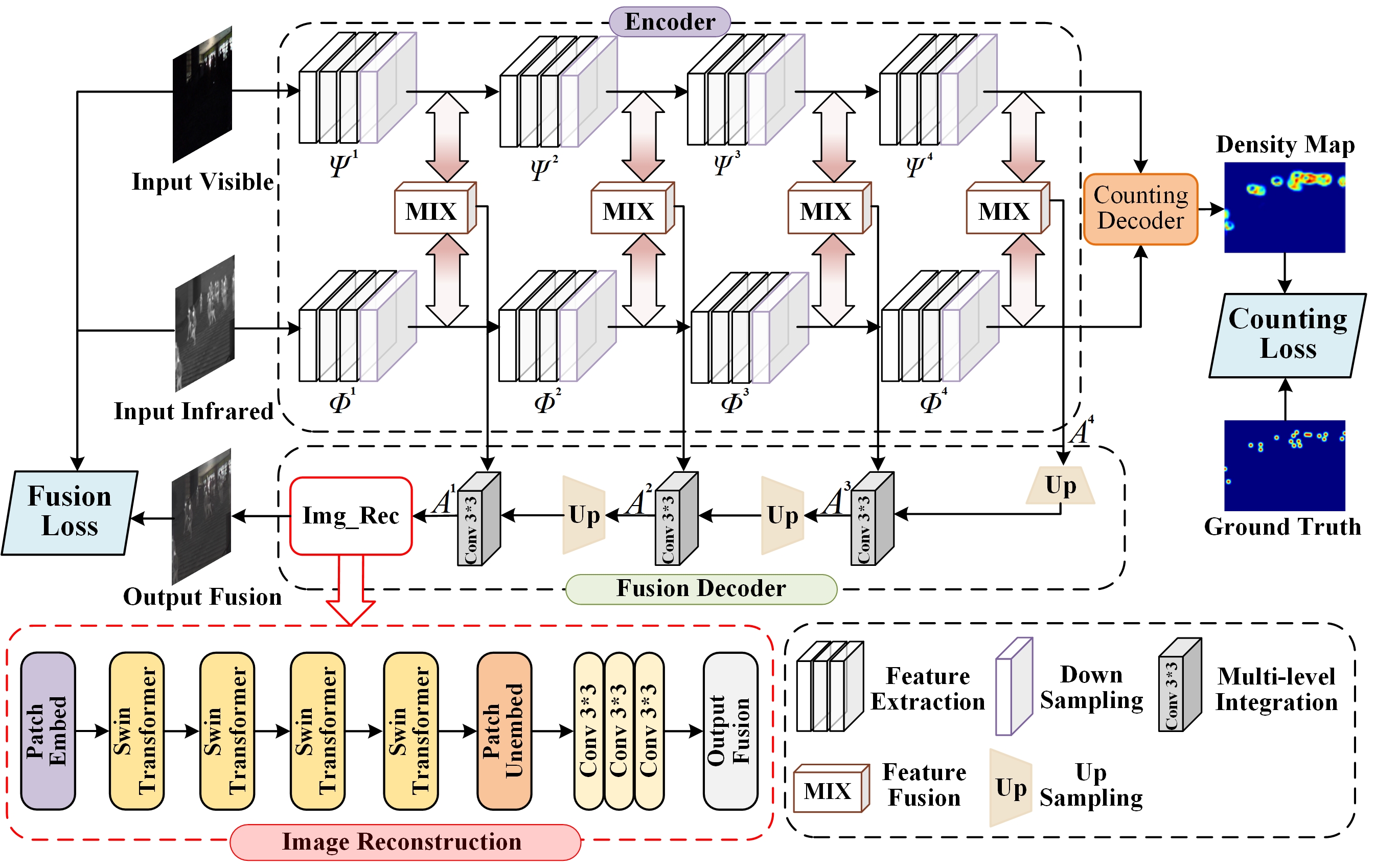}
	\caption{The overall framework of the proposed \textbf{FusionCounting}. In the encoder, visible and infrared features are extracted by $K$-layer encoder ${\Psi}^{k}$ and ${\Phi}^{k}$, respectively. The feature interaction is carried out step by step in each layer. The crowd counting decoder generates density map, based on which the counting loss is computed. Additionally, in the fusion decoder, bilinear interpolation is used to sample and integrate the fused features of each layer.~The image reconstruction module generated final image fusion results, based on which the fusion loss is calculated.}
	\label{fig:framework}
\end{figure*}

\subsection{Application-oriented image fusion}
Before 2019, VIF methods mainly aimed to generate high-quality fused images. However, because downstream applications, e.g., object detection and semantic segmentation, are not considered in the image fusion process \cite{zhang2023visible, liu2024infrared}, these fused images may not be optimal for downstream applications. From around 2019, researchers \cite{shopovska2019deep} started considering downstream applications in the process of image fusion, and this kind of methods is usually called application-oriented VIF methods \cite{zhang2023visible}.~Since then, several downstream applications have been considered in application-oriented VIF methods, such as object detection \cite{liu2022target}, pedestrian detection \cite{zheng2024pedestrian}, and semantic segmentation \cite{tang2022image,liu2023multi,fu2024segmentation,chen2024hitfusion}.~However, to the best of our knowledge, crowd counting has not been applied in application-oriented VIF as a downstream application to guide the fusion process.~In this work, we will fill this gap.

\subsection{RGB-T crowd counting}
Crowd counting is a pedestrian-oriented scene understanding task. RGB-T crowd counting has gained attention due to the complementary nature of visible and infrared images, enabling more reliable predictions in complex surveillance environments. Most existing works focus on cross-modal feature fusion. For instance, Liu et al. \cite{liu2021cross} proposed an information aggregation and distribution module for adaptive feature fusion and introduce a challenging benchmark. Zhang et al. \cite{zhang2022spatio} and Zhou et al. \cite{zhou2022defnet,zhou2024mc3} further explored cross-modal feature aggregation using attention mechanisms to enhance counting performance. 
The work by Meng et al. \cite{meng2024multi} proposed a fusion-based method for multi-modal crowd counting by introducing an auxiliary broker modality, addressing ghosting artifacts in cross-modal fusion while achieving accurate counting with minimal parameter overhead.

Despite these advances, existing methods fuse cross-modal features without fusion supervision. However, crowd counting and its upstream image fusion task are strongly correlated, a relationship we aim to explore in this study. To our knowledge, this is the first work to integrate VIF into crowd counting.

\section{Methodology} \label{Methodology}

\subsection{Overview of the proposed method}
In this study, we propose FusionCounting, a multi-task learning framework that jointly optimizes image fusion and crowd counting to leverage their mutual benefits. Compared to semantic segmentation, crowd counting requires only point-wise annotations at pedestrian locations, significantly reducing labeling costs. FusionCounting is a general and flexible framework. In this work, we implement it based on existing RGB-T crowd counting models, enabling simultaneous learning of image fusion and crowd counting.

\subsection{Multi-task learning architecture}
In visual information processing, the VIF model shares a highly similar structure with the RGB-T crowd counting model, as both follow a two-stream architecture. Building on this, our multi-task framework introduces a shared encoder for both tasks, integrating multiple cross-modal feature fusion modules to facilitate crowd counting in a manner similar to VIF. This shared encoder effectively extracts cross-modal feature representations while reducing redundancy in the network architecture. Task-specific decoders are then used to complete the VIF and crowd counting tasks.

Since the output size of the fusion task needs to be consistent with the input image, a series of upsampling fusion operations must be performed during the decoding process. In contrast, the crowd counting task only needs to generate a density map from the feature map to estimate the number of people, so there is no need to perform an upsampling operation. Here, we use ${I}_{vi}^{k} $ and ${I}_{ir}^{k} $ to denote the $k$-layer visible and infrared encoder inputs, ${\Psi}^{k}\left(\cdot\right)$ and ${\Phi}^{k}\left(\cdot\right)$ are $k$-layer encoders for visible and infrared input, respectively. $MIX\left(\cdot\right)$ denotes the interaction between the $k$-layer features of visible and infrared input:
\begin{equation}
{m}^{k}= MIX\left ( {\Psi}^{k}\left ( {I}_{vi}^{k}\right ),{\Phi}^{k}\left ( {I}_{ir}^{k}\right )\right ),
\end{equation}
\begin{equation}
\begin{aligned}
{I}_{vi}^{k}={\Psi}^{k-1}\left ( {I}_{vi}^{k-1}\right )\oplus {m}^{k-1},\\{I}_{ir}^{k}={\Phi}^{k-1}\left ( {I}_{ir}^{k-1}\right )\oplus {m}^{k-1},
\end{aligned}
\end{equation}
where ${m}^{k}$ is the result of cross-modal interaction at $k$-th layer. We use $up\left ( \cdotp \right )$ to denote a linear upsampling operation.~
The decoding result of the $k$-th layer of fusion decoder is:
\begin{equation}
{\mathcal{A}}^{k}= {Conv}_{3\times3}\left ( up\left ( {\mathcal{A}}^{k+1}\right )\oplus{m}^{k} \right ).
\label{eq33}
\end{equation}
Note that a supplement to Eq.~(\ref{eq33}) is ${\mathcal{A}}^{4}={m}^{4}$. 

Use ${\mathcal{F}}_{Count}$ and ${\mathcal{F}}_{Fusion}$ to denote decoders for the two tasks, then
\begin{equation}
\begin{aligned}
{O}_{count}= {\mathcal{F}}_{Count}\left ({I}_{vi}^{4},{I}_{ir}^{4}\right ),\\{O}_{fuison}= {\mathcal{F}}_{Fusion}\left ({\mathcal{A}}^{1}\right ),
\end{aligned}
\label{eqoo}
\end{equation}
where ${O}_{count}$ and ${O}_{fuison}$ are crowd counting results and the fusion results, respectively.

As shown in Fig.~\ref{fig:framework}, we use the encoder of an existing RGB-T crowd counting model as the shared encoder in FusionCounting. Notably, this encoder can be replaced with other crowd counting model architectures, as demonstrated in our experiments. For task-specific decoders, we retain the original regression part for crowd counting. For image fusion, we leverage hierarchical inter-layer features as decoder inputs to enable multi-scale feature representation. Since the fusion output matches the input size, upsampling layers gradually refine the inter-layer features before passing them to the image reconstruction module \cite{ma2022swinfusion} to generate fused images.

\subsection{Loss function}
The loss function of the proposed FusionCounting consists of two parts, i.e., image fusion loss ${\mathcal{L}}_{f}$ and crowd counting loss ${\mathcal{L}}_{c}$, and is defined as:${\mathcal{L}}_{overall}={\lambda }_{1}{\mathcal{L}}_{f}+{\lambda }_{2}{\mathcal{L}}_{c}$,
where $\lambda_1$ and $\lambda_2$ are weighting parameters.

\subsubsection{Image fusion loss}
The image fusion loss ${\mathcal{L}}_{f}$ is implemented with a combination of the  $\mathcal{L}_{SSIM}$,  $\mathcal{L}_{int}$,  $\mathcal{L}_{int}$ , $\mathcal{L}_{MSE}$, as is expressed:
\begin{equation}
 \mathcal{L}_f={\eta }_{1}{\mathcal{L}}_{SSIM}+{\eta }_{2}{\mathcal{L}}_{text}+{\eta }_{3}{\mathcal{L}}_{int}+{\eta }_{4}{\mathcal{L}}_{MSE},
 \label{eq:lf}
\end{equation}
where ${\eta }_{1},{\eta }_{2},{\eta }_{3},{\eta }_{4}$ are weighting parameters.

To preserve the quality of the images and measure the structural similarity between images, we use the structural similarity loss SSIM:
\begin{equation}
\begin{small}
 {\mathcal{L}}_{SSIM}=1- \frac{1}{2}\left( {SSIM}\left(I_{RGB},I_{T}\right)+{SSIM}\left(I_{RGB},I_{T}\right)\right),
\end{small}
\end{equation}
where $F$ is the generated image, $I_{RGB}$ and $I_{T}$ are visible and infrared images, respectively. To ensure that the local texture pattern of the generated image is consistent with the target image, the texture loss is also introduced:
\begin{equation}
{\mathcal{L}}_{text}= \frac{1}{HW}{\left \| \left | \nabla {I}_{F}\right |-\max \left ( \left | \nabla {I}_{RGB}\right |,\left | \nabla {I}_{T}\right |\right ) \right \|}_{1},
\end{equation}
where $\nabla$ is the Sobel gradient operator and $\left \|\cdotp \right \|$ denotes the L1 norm. To improve the accuracy of overall brightness and contrast, we also use the intensity loss for image fusion task:
\begin{equation}
{\mathcal{L}}_{int}= \frac{1}{HW}{\left \| {I}_{F}-\max \left({I}_{RGB},{I}_{T}\right)\right \|}_{1}.
\end{equation}

To accurately restore the pixel-level information of the image, we use pixel loss and calculate MSE:
\begin{equation}
  {\mathcal{L}}_{MSE}= \frac{1}{2}\left ( {\left \|{I}_{F}-{I}_{RGB} \right \|}_{2}^{2}+{\left \|{I}_{F}-{I}_{T} \right \|}_{2}^{2}\right).
\end{equation}

\subsubsection{Crowd counting loss}
The crowd counting loss item $\mathcal{L}_c$ is implemented using the commonly-used Baysiean loss item \cite{liu2021cross} as:
\begin{equation}
  {\mathcal{L}}_{c}= \displaystyle\sum_{n=1}^{N}DF\left (1- E\left [ {c}_{n}\right ]\right),
\end{equation}
where ${c}_{n}$ is the total count associated with a given label and $E\left [ {c}_{n}\right ]$ is the expectation for ${c}_{n}$. $DF\left(\cdot\right)$ represents a distance
function, which is ${l}_{1}$ distance in the experiments.

\subsection{Dynamic weighting strategy for loss functions}
To enhance learning across multiple tasks, we propose a dynamic weighting strategy to balance different loss terms, inspired by \cite{zhou2022domain}. Specifically, in a multi-task learning framework with $K$ tasks, each task's loss function is assigned a weighted coefficient $\lambda_k$. In FusionCounting, we set $K=2$. The core idea of our dynamic weighting strategy is to accelerate the convergence of the faster-learning task, identified by greater loss variation, which reflects a higher convergence rate.

While existing approaches often reduce the weighted factor of faster-learning tasks to shift focus to other tasks, we adopt the principle of the Matthew Effect \cite{gao2023alleviating}. Specifically, we amplify the weighted coefficient of the faster-learning task, enabling it to converge first and simultaneously stimulate the learning of other tasks. Thus, the weighted coefficient $\lambda_k\left(t\right)$ of the $k$-th task in $t$-th epoch could be expressed as:
\begin{equation}
 \begin{aligned}
    {\lambda }_{k}\left ( t\right )= \frac{K}{K-1}&\left ( 1-\frac{exp\left ( {w}_{k}\left ( t-1\right )*Z\right )}{\sum_i exp\left ( {w}_{i}\left ( t-1\right )*Z\right )}\right )\\
 {w}_{k}\left ( t-1\right )&= \frac{\mathcal{L}_{k}\left ( t-1\right )}{\mathcal{L}_{k}\left ( t-2\right )},
 \end{aligned}
\end{equation}
where $w_k\left(t-1\right)$ represents the relative descending rate of $t-1$ epoch of loss values, $\mathcal{L}_k\left(t-1\right)$ and $\mathcal{L}_k\left(t-2\right)$ represent the $\left(t-1\right)$-th loss value and $\left(t-2\right)$-th loss value of the $k$-th task, $Z$ represents a factor to realize soft weighting. Similar to \cite{hinton2015distilling}, $\frac{K}{K-1}$ is set to ensure that $\sum_i\lambda\left(t\right)=K$. In the detailed implementation, $w_k\left(t\right)$ is set to 1, when $t$ equals to 1 and 2; the parameter $Z$ is set to 3.

The above multi-task training algorithm with the developed dynamic weighting strategy is provided  in Algorithm \ref{alg:algorithm1} 

\begin{algorithm} 
	\caption{Multi-task training process}     
	 \label{alg:algorithm1}       
	\begin{algorithmic}[1] 
	\Require N pairs of RGBT images ${X}^{N}$  correspond to real      density maps ${D}_{{X}^{N}}^{DT}$;   
    \Ensure  Multi-task learning model network parameters $\varTheta$ ;  
    \State Read the data and labels of training set, verification set and test set, and conduct data preprocessing;    
    \For {epoch=1 to 200}
       \For {i=0 to N}
   \State Input ${X}^{i}$ into the model and output the estimated 
   \State density map ${D}_{{X}^{i}}^{ET}$ with the fused image ${F}^{i}$; 
   \State loss ${\mathcal{L}}_{c}$ of count was calculated by ${D}_{{X}^{i}}^{ET}$ and \State${D}_{{X}^{i}}^{DT}$, and fusion loss ${\mathcal{L}}_{f}$ was calculated by ${F}^{i}$ \State and ${X}^{i}$;
   \State And then we calculate ${\mathcal{L}}_{t}$=${\lambda }_{1}*{\mathcal{L}}_{c}$+${\lambda }_{2}* {\mathcal{L}}_{f}$;
   \State back-propagate ${\mathcal{L}}_{t}$ to update $\varTheta$ ;
   
   \EndFor
   \EndFor
   \State\Return return the optimal parameters $\varTheta$ 
   \end{algorithmic} 
   \label{algorithm}
\end{algorithm} 

\subsection{Adversarial training scheme}

To improve robustness, we develop an adversarial training scheme for the proposed method. This scheme integrates samples generated by PGD attacks \protect\cite{waghela2024robust} into the training set, allowing the model to learn and adapt to such attack patterns. As a result, it strengthens the model’s defense against malicious attacks and enhances overall robustness. The adversarial training algorithm is detailed in Algorithm \ref{algorithm}.
\begin{algorithm} [h]
	\caption{Adversarial training algorithm} 
	 \label{1} 
	\begin{algorithmic}[1] 
	\Require N pairs of clean RGBT images ${X}^{N}$  correspond to real density maps ${D}_{{X}^{N}}^{DT}$; 
 \Ensure  Adversarial model network parameters $\varTheta_{adv}$ ;  
 \State Read clean data ${X}_{c}^{N}$ and labels , ${X}_{c}^{N}\xrightarrow[]{PGD}{X}_{adv}^{N}$; 
 \For {epoch=1 to 200}
 \For {i=0 to 2*N}
 \State Input ${X}^{i}$ into the model and output the estimated 
 \State density map ${D}_{{X}^{i}}^{ET}$ with the fused image ${F}^{i}$; 
 \State loss ${\mathcal{L}}_{c}$ of count was calculated by ${D}_{{X}^{i}}^{ET}$ and \State${D}_{{X}^{i}}^{DT}$, and fusion loss ${\mathcal{L}}_{f}$ was calculated by ${F}^{i}$ \State and ${X}_{c}^{N}$;
 \State And then we calculate ${\mathcal{L}}_{t}$=${\lambda }_{1}*{\mathcal{L}}_{c}$+${\lambda }_{2}* {\mathcal{L}}_{f}$
 \State back-propagate ${\mathcal{L}}_{t}$ to update $\varTheta_{adv}$ ;
 
 \EndFor
 \EndFor
 \State\Return return the optimal parameters $\varTheta_{adv}$ 
 \end{algorithmic} 
 \label{algorithm}
\end{algorithm}

\subsection{Model inference}
In the inference stage, image fusion results are provided by the fusion decoder, while crowd counting results are provided by the counting decoder.~This is different from most existing application-oriented VIF methods (e.g., SeAFusion \cite{tang2022image} and SegMiF \cite{liu2023multi}), where the results of downstream applications, such as segmentation and detection, are obtained by running another segmentation or detection model on the fused images.

\renewcommand{\algorithmicrequire}{\textbf{Input:}}
\renewcommand{\algorithmicensure}{\textbf{Output:}}

\section{Experiments} \label{exp}

\subsection{Experimental settings}
\subsubsection{Implementation details}
Our FusionCounting is implemented in Pytorch on a 
NVIDIA GeForce RTX 4090 GPU. Adam optimizer is used to update parameters and cosine annealing algorithm is used to adjust the learning rate to ensure the stability and convergence speed of the training process. The learning rate is 1$\times$10$^{-5}$.~${\eta }_{1},{\eta }_{2},{\eta }_{3},{\eta }_{4}$ in Eq.~(\ref{eq:lf}) are 20, 30, 30, and 100, respectively. In addition, since there is no universally fixed configuration for PGD, we follow the principles outlined in the paper to balance attack strength and imperceptibility. Specifically, the maximum perturbation parameter $\epsilon$, step size $\alpha$, and number of iterations are set to 20, 5, and 7, respectively.

\subsubsection{Datasets}
We evaluate the developed FusionCounting on the following two widely-used datasets: RGBT-CC \cite{liu2021cross} and DroneRGBT \cite{Peng2020Dronergbt}. RGBT-CC is a public cross-modal visible-infrared crowd counting dataset captured from a surveillance perspective. Since its original samples are not strictly aligned, this study selected a subset with aligned data to meet the requirements of the joint image fusion task, consisting of 176 RGB-T image pairs for training, 23 for validation, and 109 for testing. Compared to the original test set with an average crowd count of 74.12, our selected subset has a significantly higher average of 130.6, indicating denser and more challenging scenes that may lead to performance drops. 
DroneRGBT is a visible-infrared crowd dataset captured from a drone perspective, consisting of 3,600 image pairs with a 50\% training and 50\% test split. To monitor overfitting, we used 10\% of the training set as a validation set, which led to a slight decrease in training performance.

\subsubsection{Evaluation metrics}
\textbf{VIF metrics}.~For the image fusion task, eight commonly-used evaluation metrics were selected: 
metrics based on information entropy (PSNR \cite{jagalingam2015review} and Qabf \cite{xydeas2000objective}), metrics based on correlation (CC \cite{shah2013multifocus} and SCD \cite{aslantas2015new}), metrics based on structural similarity (SSIM \cite{wang2004image}) and image features metrics (AG \cite{cui2015detail}, SF \cite{cui2015detail}, SD \cite{aslantas2015new}). These evaluation metrics comprehensively evaluate the quality of fused images from multiple aspects, ensuring the objectivity and reliability of the evaluation results.

\noindent\textbf{Crowd counting metrics}.~Following existing studies \cite{kong2024multiscale}, we use the classical Root Mean Square Error (RMSE) and the Grid Average Mean absolute Error (GAME) \cite{Guerrero2015GAME} as evaluation metrics. For the GAME calculation, the density map is divided into non-overlapping $4^{l}$ regions, and the error is calculated separately in each region. Lower GAME($l$) values represent better crowd counting performance.

\subsubsection{Compared methods}

\noindent\textbf{VIF methods}.~We selected nine state-of-the-art VIF methods for comparison, i.e., RFN-Nest \cite{li2021rfn}, SwinFusion \cite{ma2022swinfusion}, Rse2Fusion \cite{wang2022res2fusion}, UNFusion \cite{wang2021unfusion}, PSCM \cite{qi2024ps}, FCDFusion \cite{li2025fcdfusion}, SeAFusion  \cite{tang2022image}, PSFusion\cite{tang2023rethinking}, MRFS \cite{zhang2024mrfs}.~In these methods, SeAFusion, PSFusion and MRFS are application-oriented VIF methods.

\noindent\textbf{Crowd counting methods}.~We compared our method with six crowd counting methods, i.e., IADM \cite{liu2021cross}, CSCA \cite{zhang2022spatio}, MC$^3$Net \cite{zhou2024mc3}, DEFNet \cite{zhou2022defnet}, CAG \cite{yang2024cagnet}, DLF-IA \cite{cheng2024late}.~These methods take RGB-T image pairs as input. 

\begin{figure*}[!h]
 \centering
 \includegraphics[width=1\textwidth]{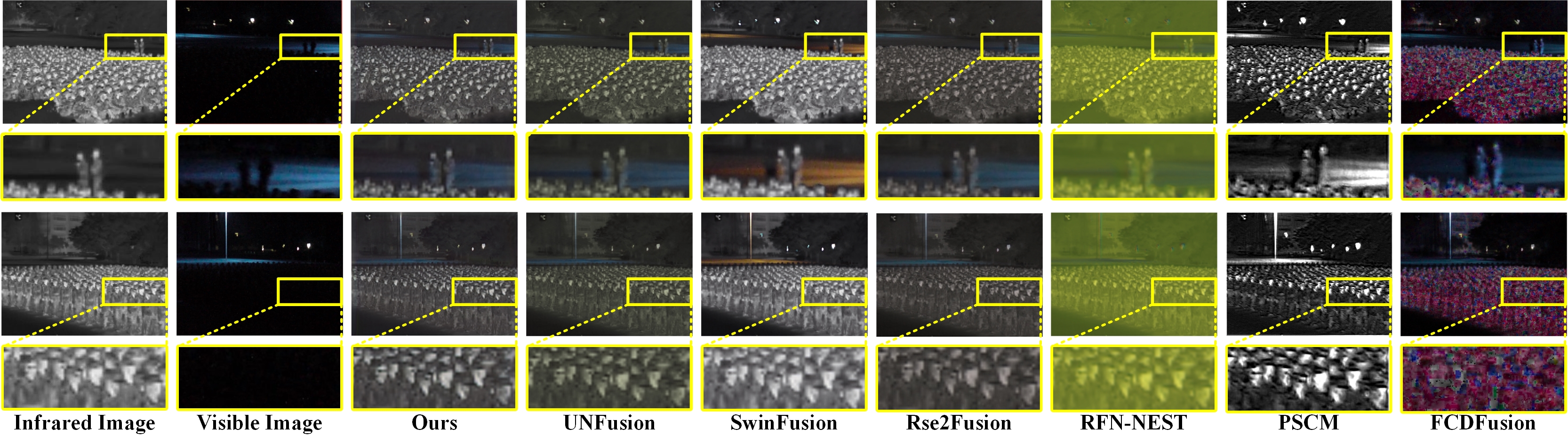}
 \caption{Qualitative comparison of image fusion performance between the proposed FusionCounting and the compared VIF methods on RGBT-CC. Compared with other methods, our FusionCounting achieves a clear fused image with less color distortion. Here, FusionCounting is implemented based on MC$^3$Net, which is the same in Figure \ref{fig:counting-qualitative}. }
 \label{fig:fusion-qualitative}
\end{figure*}

\begin{figure*}[!h]
 \centering
 \includegraphics[width=1\textwidth]{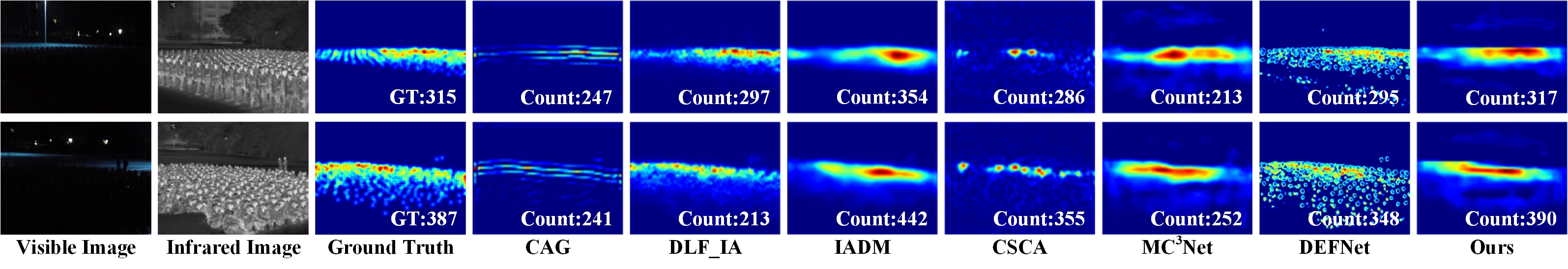}
 \caption{Qualitative comparison of crowd counting performance between the proposed FusionCounting and the compared crowd counting methods on RGBT-CC. It can be observed that the counting results generated by ours are closer to ground truth than other methods.}
 \label{fig:counting-qualitative}
\end{figure*}

\subsection{Results of visible-infrared image fusion}

\noindent\textbf{Qualitative results}.~The qualitative comparison of image fusion performance is shown in Fig.~\ref{fig:fusion-qualitative}. 
As can be seen, most methods, including 
FCDFusion, RFNFusion, SwinFusion, UNFusion, and PSCM, 
have significant 
color distortion in the fused image.~Though the fused image of Res2Fusion keeps color information in a good way, the fused results present more blurring than ours.

 \noindent\textbf{Quantitative results}.~The quantitative image fusion results 
are shown in Table \ref{fig:fusion-performance}. We implemented our FusionCounting based on several crowd counting methods, i.e., 
IADM \cite{liu2021cross}, CSCA \cite{zhang2022spatio}, MC$^3$Net \cite{zhou2024mc3}, and DEFNet \cite{zhou2022defnet}. The results show that FusionCounting achieves highly competitive performance across several metrics. This is facilitated by the mutual reinforcement of the two types of supervision, which reflects the advantages of multi-task supervision. For example, when IADM is integrated into our FusionCounting framework, the best SSIM and PSNR results are obtained.
It is worth noting that our method is trained using much less RGB-T image pairs than the compared VIF methods.

\begin{table}[t]

\caption{Quantitative comparison of image fusion performance on the RGBT-CC dataset. Results in \textcolor{red}{red}, \textcolor{blue}{blue} and \textcolor{darkgreen}{green} represent the best, the second-best and the third-best.}
\resizebox{1\columnwidth}{!}{
\begin{tabular}{ccccccccc}
\toprule
Methods & Qabf$\uparrow$ & SSIM$\uparrow$ & PSNR$\uparrow$ & CC$\uparrow$ & SCD$\uparrow$ & SD$\uparrow$  & AG$\uparrow$ & SF$\uparrow$ \\ 
\midrule
RFN-Nest  & \textcolor{darkgreen}{0.57} & 0.64 & 13.86 & 0.44 & 0.72 & 40.37 & 3.63 & \textcolor{darkgreen}{11.43} \\
UNFusion  & 0.54 & \cellcolor{lightblue}\textcolor{blue}{0.67} & 14.75 & \cellcolor{lightblue}\textcolor{blue}{0.58} & 1.49 & 27.67  & 2.64  & 7.22  \\
Rse2Fusion  & 0.52 & \cellcolor{lightblue}\textcolor{blue}{0.67} & 14.70 &\cellcolor{lightblue}\textcolor{blue}{0.58} & 1.46 & 26.60  & 2.58 & 7.08  \\
SwinFusion  &  \cellcolor{lightred}{\color[HTML]{FF0000} 0.67} & 0.65 & 13.47 & \textcolor{darkgreen}{0.56} & \cellcolor{lightred}{\color[HTML]{FF0000} 1.73} & \cellcolor{lightblue}\textcolor{blue}{44.98}  & \cellcolor{lightblue}\textcolor{blue}{4.11} & \cellcolor{lightblue}\textcolor{blue}{11.73} \\
PSCM  & 0.20 & 0.40 & \cellcolor{lightblue}\textcolor{blue}{56.87} & 0.53 & 1.44 & \cellcolor{lightred}{\color[HTML]{FF0000} 104.89} & \cellcolor{lightred}{\color[HTML]{FF0000} 9.57} & \cellcolor{lightred}{\color[HTML]{FF0000} 25.23} \\
FCDFusion   & 0.30 & 0.18 & 10.78 & \cellcolor{lightred}{\color[HTML]{FF0000} 0.76} & 1.47 & \textcolor{darkgreen}{43.01}  & 3.52 & 10.60 \\ 
\hline 
IADM + FC  & 0.55 & \cellcolor{lightred}{\color[HTML]{FF0000} 0.73} & \cellcolor{lightred}{\color[HTML]{FF0000} 63.98}  & 0.55 & \cellcolor{lightblue}\textcolor{blue}{1.62} & 38.42  & 3.68 & 10.48 \\
CSCA + FC  & 0.52 & 0.65 & 15.42 & \cellcolor{lightblue}\textcolor{blue}{0.58} & 1.58 & 35.24  & \textcolor{darkgreen}{3.81} & 10.06 \\
MC$^3$Net + FC   & 0.39 & \textcolor{darkgreen}{0.66} & 15.63 & 0.52  & 1.22 & 33.52  & 3.22 & 9.48  \\
DEFNet + FC  & \cellcolor{lightblue}\textcolor{blue}{0.61} & 0.65 & {\color[HTML]{75BD42} 16.01} & \cellcolor{lightblue}\textcolor{blue}{0.58} & \textcolor{darkgreen}{1.59} & 31.75  & 3.64 & 10.74 \\ 
\bottomrule
\end{tabular}
}
\label{fig:fusion-performance}
\end{table}

\subsection{Results of RGB-T crowd counting}
\subsubsection{Comparison with RGB-T crowd counting methods}

 \noindent\textbf{Qualitative results}.~Qualitative comparison of crowd counting results is shown in Fig.~\ref{fig:counting-qualitative}.~As can be seen, 
both predicted density map and counting results given by FusionCounting are very close to the ground truth.~In contrast, the compared methods show larger difference with the ground-truth counting result. This is attributed to the benefit of the VIF task on the crowd counting task in our FusionCounting.

\begin{table}[t]
\caption{Quantitative comparison of crowding counting performance on the RGBT-CC dataset.} 
\resizebox{1\columnwidth}{!}{
\begin{tabular}{cccccc}

\hline
{Methods}&GAME(0)$\downarrow$ & GAME(1)$\downarrow$ & GAME(2)$\downarrow$ & GAME(3)$\downarrow$ & RMSE$\downarrow$ \\ \hline
 CAG& 31.38 & 40.28 & 51.04 & 63.52 & 56.18 \\
 DLF-IA & \cellcolor{lightblue}\textcolor{blue}{19.95} & \cellcolor{lightblue}\textcolor{blue}{26.74} & 36.03 & 50.12 & 35.74 \\ 
IADM & 27.64 & 39.65 & 48.08 & 61.47 & 42.34 \\
 CSCA& 24.17 & 33.81 & 38.29 & 50.09 & 35.4 \\
  MC$^3$Net & 21.30 & 30.99 & 41.22 & 56.6 & \textcolor{darkgreen}{ 32.48} \\
 DEFNet &23.35 & 36.97 & 43.3 & 53.54 & 37.89 \\ 
\hline 
IADM + FC  & \textcolor{darkgreen}{ 20.49} & 29.27 & \cellcolor{lightblue}\textcolor{blue}{35.77} & \textcolor{darkgreen}{ 47.95} & \cellcolor{lightblue}\textcolor{blue}{30.22} \\
 CSCA + FC   & 21.23 & \textcolor{darkgreen}{ 29.20}  & \textcolor{darkgreen}{ 36.00} & \cellcolor{lightblue}\textcolor{blue}{47.34} & 35.23 \\
  MC$^3$Net + FC & \cellcolor{lightred}{\color[HTML]{FF0000} 15.84} & \cellcolor{lightred}{\color[HTML]{FF0000} 23.96} & \cellcolor{lightred}{\color[HTML]{FF0000} 32.5}  &\cellcolor{lightred} {\color[HTML]{FF0000} 45.56} &\cellcolor{lightred} {\color[HTML]{FF0000} 26.02} \\
 DEFNet + FC& 21.50 & 36.46 & 42.64 & 53.47 & 36.14 \\ \hline
\end{tabular}
}

\label{tablecount}
\end{table}

 \noindent\textbf{Quantitative results}.~\mbox{Table \ref{tablecount}} reports the quantitative comparison of our FusionCounting with the recent representative RGB-T crowd counting methods on RGBT-CC.~It can be clearly seen that our FusionCounting shows significant competitive advantages over various metrics for the crowd counting task. More importantly, compared with the basic version of IADM, CSCA, MC$^3$Net and DEFNet, the proposed FusionCounting built on these methods 
achieves significant improvements. Although DEFNet+FC appears less competitive than other FC-based models, it still clearly outperforms the original DEFNet and several classical baselines, demonstrating the robustness of our FC module. Moreover, compared with 
CAG and DLF-IA methods, our FusionCounting also shows an 
improvement in a large margin.~These results further demonstrate the benefits of our multi-task learning framework on the crowd counting task.

\subsubsection{Compared with application-oriented VIF methods}
We compared our FusionCounting 
with three representative 
application-oriented VIF methods: SeAFusion \cite{tang2022image}, PSFusion \cite{tang2023rethinking}, and MRFS \cite{zhang2024mrfs}. First, we applied these methods to the RGBT-CC dataset to generate fused images, then used MC$^3$Net for crowd counting. The results, shown in Table~\ref{table:count-appVIF}, indicate that FusionCounting outperforms these three methods. This is because these 
methods were jointly trained for image fusion and semantic segmentation, making them less effective at learning features beneficial for crowd counting. In contrast, FusionCounting directly learns crowd counting-relevant features through joint training of VIF and crowd counting. These findings suggest that features learned via semantic segmentation do not generalize well to crowd counting, highlighting the necessity of jointly training VIF with crowd counting to obtain good crowd counting performance.

\begin{table}[t]
\begin{scriptsize}

\caption{Quantitative comparison of our FusionCounting and three representative segmentation-oriented VIF methods on the RGBT-CC dataset. Here, the FusionCounting is implemented based on MC$^3$Net.}
\begin{footnotesize}
\resizebox{1\columnwidth}{!}{
\begin{tabular}{ccccc}
\hline
Method&SeAFusion&PSFusion&MRFS&\textbf{FusionCounting}\\ \hline 
Year&2022&2023&2024&Ours\\\hline
GAME(0)$\downarrow$&27.89&28.90&34.21& \cellcolor{lightred}\textbf{15.84}\\
GAME(1)$\downarrow$&37.29&39.06&41.76& \cellcolor{lightred}\textbf{23.96}\\
GAME(2)$\downarrow$&46.70&48.96&50.23& \cellcolor{lightred}\textbf{32.50}\\
GAME(3)$\downarrow$&60.24&63.11&62.59& \cellcolor{lightred}\textbf{45.56}\\
RMSE$\downarrow$&44.14&45.01&51.12& \cellcolor{lightred}\textbf{26.02}\\\hline
\end{tabular}
}
\end{footnotesize}
\label{table:count-appVIF}
\end{scriptsize}

\end{table}

\begin{table}[t]
\centering
\caption{Model complexity. (model+FC)/original model is shown.}
\resizebox{0.5\textwidth}{!}{
\begin{tabular}{lcccc}
\hline
Methods    & CSCA         & DEFNet         & IADM          & MC$^3$Net       \\ \hline
FLOPs (G)  & 163.1 / 94.06 & 563.87 / 494.27 & 33.07 / 22.13 & 805.05 / 713.99 \\
Params (M) & 20.57 / 17.19 & 45.57 / 45.16   & 27.36 / 25.67 & 268.08 / 260.52 \\ \hline
\end{tabular}
}
\label{flps}
\end{table}


\begin{table*}[!t]

\caption{Effectiveness of multi-task learning and dynamic weighting 
of our FusionCounting on the RGBT-CC dataset. Here, `ST' represents the model only with a single task, `w/o DW' represents the model of FusionCounting without dynamic weighting factors, and `Ours' represents the original FusionCounting. \textbf{Bold} denotes the optimal.}
\centering
\resizebox{1\textwidth}{!}{
\begin{tabular}{lc|cccccccc|ccccc}
\hline
\multirow{2}{*}{Baselines} & \multirow{2}{*}{Settings} & \multicolumn{8}{c|}{Visible-infrared image fusion} & \multicolumn{5}{c}{Crowd counting} \\ \cline{3-15} 
& & Qabf$\uparrow$ & SSIM$\uparrow$ & PSNR$\uparrow$ & CC$\uparrow$ & SCD$\uparrow$ & SD$\uparrow$ & AG$\uparrow$ & SF$\uparrow$ & GAME(0)$\downarrow$ & GAME(1)$\downarrow$ & GAME(2)$\downarrow$ & GAME(3)$\downarrow$ & RMSE$\downarrow$ \\ \hline
\multirow{3}{*}{IADM} & ST & \cellcolor{lightred}\textbf{0.55} & \cellcolor{lightred}\textbf{0.73} & 63.29 & \cellcolor{lightred}\textbf{0.55} & 1.60 & 37.69 & 3.63 & 10.35 & 27.64 & 39.65 & 48.08 & 61.47 & 42.34 \\
& w/o DW & 0.44 & 0.70 & 63.63 & 0.54 & 1.45 & 34.93 & 3.39 & 9.94 & 22.99 & 33.42 & 40.20 & 52.95 & 35.11 \\ 
& \textbf{Ours} & \cellcolor{lightred}\textbf{0.55} & \cellcolor{lightred}\textbf{0.73} & \cellcolor{lightred}\textbf{63.98} & \cellcolor{lightred}\textbf{0.55} & \cellcolor{lightred}\textbf{1.62} & \cellcolor{lightred}\textbf{38.42} & \cellcolor{lightred}\textbf{3.68} & \cellcolor{lightred}\textbf{10.48} & \cellcolor{lightred}\textbf{20.49} & \cellcolor{lightred}\textbf{29.27} & \cellcolor{lightred}\textbf{35.77} & \cellcolor{lightred}\textbf{47.95} & \cellcolor{lightred}\textbf{30.22} \\  \Xhline{1pt} 

\multirow{3}{*}{CSCA} & ST & 0.43 & 0.63 & 13.73 & 0.35 & 0.49 & 34.95 & \cellcolor{lightred}\textbf{3.85} & \cellcolor{lightred}\textbf{11.11} & 24.17 & 33.81 & 38.29 & 50.09 & 35.40 \\
& w/o DW & 0.31 & 0.61 & 13.69 & 0.33 & 0.43 & 31.27 & 3.32 & 8.64 & 24.23 & 33.20 & 38.92 & 50.36 & 38.24 \\ 
& \textbf{Ours} & \cellcolor{lightred}\textbf{0.52} & \cellcolor{lightred}\textbf{0.65} & \cellcolor{lightred}\textbf{15.42} & \cellcolor{lightred}\textbf{0.58} & \cellcolor{lightred}\textbf{1.58} & \cellcolor{lightred}\textbf{35.24} & 3.81 & 10.06 & \cellcolor{lightred}\textbf{21.23} & \cellcolor{lightred}\textbf{29.20} & \cellcolor{lightred}\textbf{36.00} & \cellcolor{lightred}\textbf{47.34} & \cellcolor{lightred}\textbf{35.23} \\\Xhline{1pt} 

\multirow{3}{*}{MC$^3$Net} & ST & \cellcolor{lightred}\textbf{0.55} & \cellcolor{lightred}\textbf{0.69} & 15.60 & 0.33 & 0.43 & \cellcolor{lightred}\textbf{34.79} & \cellcolor{lightred}\textbf{3.67} & \cellcolor{lightred}\textbf{10.23} & 21.30 & 30.99 & 41.22 & 56.60 & 32.48 \\
& w/o DW & 0.36 & 0.66 & 15.15 & 0.49 & 1.12 & 29.41 & 3.10 & 8.42 & 23.67 & 32.10 & 40.86 & 56.82 & 39.31 \\ 
& \textbf{Ours} & 0.39 & 0.66 & \cellcolor{lightred}\textbf{15.63} & \cellcolor{lightred}\textbf{0.52} & \cellcolor{lightred}\textbf{1.22} & 33.52 & 3.22 & 9.48 & \cellcolor{lightred}\textbf{15.84} & \cellcolor{lightred} \textbf{23.96} & \cellcolor{lightred}\textbf{32.50} & \cellcolor{lightred}\textbf{45.56} & \cellcolor{lightred}\textbf{26.02} \\ \Xhline{1pt} 

\multirow{3}{*}{DEFNet} & ST & 0.59 & \cellcolor{lightred}\textbf{0.66} & 15.76 & \cellcolor{lightred}\textbf{0.59} & 1.45 & 27.28 & 3.04 & 9.03 & 23.35 & 36.97 & 43.30 & 53.54 & 37.89 \\
& w/o DW & 0.42 & 0.60 & 13.55 & 0.31 & 0.32 & 30.65 & 3.58 & 10.61 & 71.70 & 79.26 & 88.32 & 97.41 & 116.31 \\ 
& \textbf{Ours} & \cellcolor{lightred}\textbf{0.61} & 0.65 & \cellcolor{lightred}\textbf{16.01} & 0.58& \cellcolor{lightred}\textbf{1.59} & \cellcolor{lightred}\textbf{31.75} & \cellcolor{lightred}\textbf{3.64} & \cellcolor{lightred}\textbf{10.74} & \cellcolor{lightred}\textbf{21.50} & \cellcolor{lightred}\textbf{36.46} & \cellcolor{lightred}\textbf{42.64} & \cellcolor{lightred}\textbf{53.47} & \cellcolor{lightred}\textbf{36.14} \\\hline
\end{tabular}
}
\label{table:ablation-rgbtcc-12}
\end{table*}

\subsubsection{Comparison of parameters}
Table \ref{flps} reports the model complexity of different baselines with and without the proposed FusionCounting module. The results show that adding FusionCounting moderately increases FLOPs and parameter counts across all methods, while maintaining a reasonable computational overhead.

\subsection{Ablation studies}

\begin{table*}[!t]
\centering
\caption{Results of ablation studies on the designs of multi-task learning and dynamic weighting factors of our FusionCounting on DroneRGBT dataset. Here, ``ST'' represents the model only with a single task, ``w/o DW'' represents the model of FusionCounting without the dynamic weighting factors, and ``Ours'' represents the original FusionCounting. \textbf{Bold} denotes the optimal.}
\resizebox{1\textwidth}{!}{
\begin{tabular}{lc|cccccccc|ccccc}
\hline
\multirow{2}{*}{Baseline} & \multirow{2}{*}{Setting} & \multicolumn{8}{c|}{Visible-infrared image fusion} & \multicolumn{5}{c}{Crowd counting} \\\cline{3-15} 
& & Qabf$\uparrow$ & SSIM$\uparrow$ & PSNR$\uparrow$ & CC$\uparrow$ & SCD$\uparrow$ & SD$\uparrow$ & AG$\uparrow$ & SF$\uparrow$ & GAME(0)$\downarrow$ & GAME(1)$\downarrow$ & GAME(2)$\downarrow$ & GAME(3)$\downarrow$ & RMSE$\downarrow$ \\ \hline
\multirow{3}{*}{IADM} & ST & 0.40  & \cellcolor{lightred}\textbf{0.69}  & 62.89 & 0.65 & 1.33 & 33.50 & \cellcolor{lightred}\textbf{4.25}  & \cellcolor{lightred}\textbf{12.23}  & 11.43  & 15.08  & 19.23 & 24.58  & 17.81 \\
& w/o DW & 0.38    & \cellcolor{lightred}\textbf{0.69}   & 62.89   & 0.65   & 1.37   & 33.34  & 4.01       & 11.44  & 11.43  & \cellcolor{lightred}\textbf{14.10} & \cellcolor{lightred}\textbf{17.86} & \cellcolor{lightred}\textbf{22.71}  & 17.44 \\ 
& \textbf{Ours} & \cellcolor{lightred}\textbf{0.41}  & \cellcolor{lightred}\textbf{0.69}  & \cellcolor{lightred}\textbf{63.02} & \cellcolor{lightred}\textbf{0.65} & \cellcolor{lightred}\textbf{1.39} & \cellcolor{lightred}\textbf{33.68} & 4.12  & 12.08   & \cellcolor{lightred}\textbf{11.16}  & 14.57  & 18.97 & 23.24  & \cellcolor{lightred}\textbf{17.21} \\\Xhline{1pt}
\multirow{3}{*}{CSCA} & ST  & \cellcolor{lightred}\textbf{0.44}  & \cellcolor{lightred}\textbf{0.69}  & 15.39 & 0.67 & 1.33 & 31.52 & \cellcolor{lightred}\textbf{4.30}  & \cellcolor{lightred}\textbf{12.79}  & 12.15  & 14.92  & 18.48 & 23.59  & 19.09 \\
& w/o DW & 0.38    & \cellcolor{lightred}\textbf{0.69}   & \cellcolor{lightred}\textbf{15.60}  & 0.67   & \cellcolor{lightred}\textbf{1.38}  & \cellcolor{lightred}\textbf{31.61}  & 4.04       & 11.57 & 11.67  & 14.10 & \cellcolor{lightred}\textbf{17.66} & \cellcolor{lightred}\textbf{22.44}  & 18.20 \\ 
& \textbf{Ours}  & 0.42  & \cellcolor{lightred}\textbf{0.69}  & 15.46 & \cellcolor{lightred}\textbf{0.67} & 1.35 & 31.39 & 4.26  & 12.34  & \cellcolor{lightred}\textbf{11.09}  & \cellcolor{lightred}\textbf{14.02}  & 17.85 & 23.02  & \cellcolor{lightred}\textbf{17.75} \\\Xhline{1pt}
\multirow{3}{*}{MC$^3$Net} & ST & \cellcolor{lightred}\textbf{0.36}  & \cellcolor{lightred}\textbf{0.70}  & \cellcolor{lightred}\textbf{15.39} & 0.67 & 1.29 & 30.07  & \cellcolor{lightred}\textbf{3.78}  & 10.45 & 8.52 & \cellcolor{lightred}\textbf{10.60} & \cellcolor{lightred}\textbf{13.67} & \cellcolor{lightred}\textbf{18.74}  & 13.08  \\
& w/o DW & 0.32    & 0.68    & 15.33  & 0.66   & 1.30   & 29.91  & 3.54       & 9.72 & 9.36 & 11.70 & 14.92 & 19.97  & 14.51 \\ 
& \textbf{Ours}  & 0.35  & 0.69  & 15.36 & \cellcolor{lightred}\textbf{0.67} & \cellcolor{lightred}\textbf{1.34} & \cellcolor{lightred}\textbf{30.69} & 3.76  & \cellcolor{lightred}\textbf{10.46} & \cellcolor{lightred}\textbf{8.44} & 10.64  & 13.84 & 19.21  & \cellcolor{lightred}\textbf{12.37} \\\Xhline{1pt}
\multirow{3}{*}{DEFNet} & ST & \cellcolor{lightred}\textbf{0.47}  & \cellcolor{lightred}\textbf{0.70}  & 15.41 & 0.68 & 1.33 & 29.71 & \cellcolor{lightred}\textbf{4.33}  & \cellcolor{lightred}\textbf{12.63}   & 13.20 & 15.04  & 17.74 & 20.96  & 20.22 \\
& w/o DW & 0.37   & 0.68   & 15.45  & 0.66   & 1.36   & \cellcolor{lightred}\textbf{32.55}   & 3.98       & 11.48 & 11.67  & 14.10 & 17.40  & 20.74  & 19.56 \\ 
&  \textbf{Ours} & 0.41  & 0.69  & \cellcolor{lightred}\textbf{15.60} & \cellcolor{lightred}\textbf{0.68} & \cellcolor{lightred}\textbf{1.37} & 31.08 & 4.15  & 12.22 & \cellcolor{lightred}\textbf{11.23}  & \cellcolor{lightred}\textbf{13.20} & \cellcolor{lightred}\textbf{15.84} & \cellcolor{lightred}\textbf{18.92}  & \cellcolor{lightred}\textbf{17.38} \\\hline
\end{tabular}
}
\label{table:ablation-DroneRGBT-12}
\end{table*}

 \subsubsection{Effectiveness of multi-task learning}
 Tables \ref{table:ablation-DroneRGBT-12} and \ref{table:ablation-rgbtcc-12} present the results of ablation studies on multi-task learning on two datasets. In the table, `ST' indicates that the model is trained for a single task (either VIF or crowd counting), i.e., only fusion decoder or counting decoder is utilized. Comparing the single-task settings with the original setting (denoted as `Ours'), it can be observed that the proposed FusionCounting significantly enhances both image fusion and crowd counting results, regardless of the baseline used. This improvement is attributed to the mutual benefits between VIF and crowd counting in our FusionCounting framework.

\begin{figure}[!t]
	\centering
	\includegraphics[width=1\columnwidth]{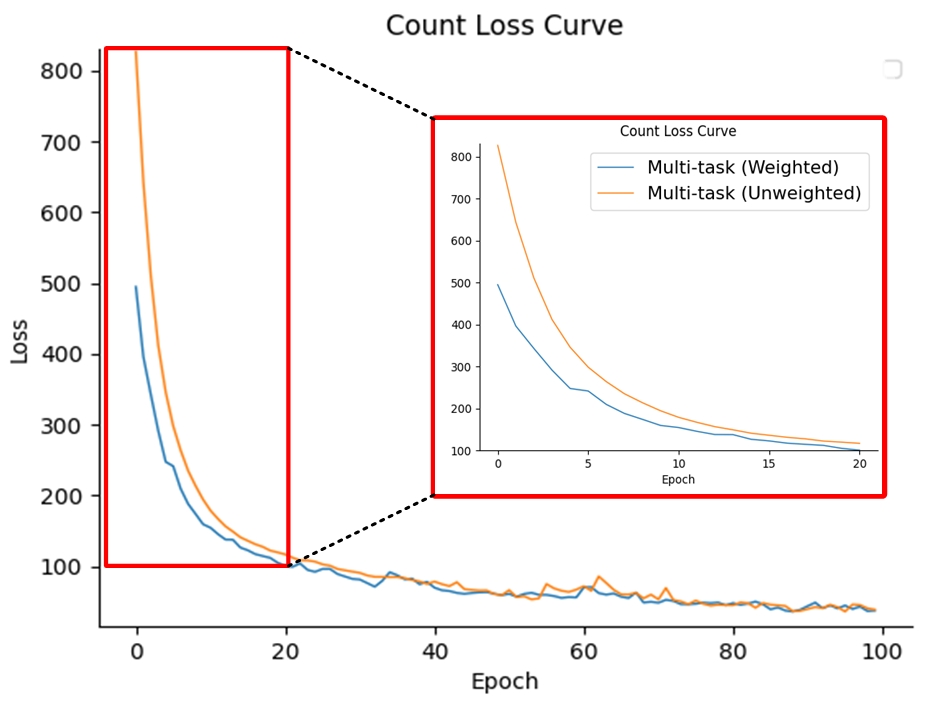}
	\caption{Visualization of training curves about the design dynamic weighting factor in the multi-task loss function.}
	\label{fig:loss}
\end{figure}

\subsubsection{Effectiveness of dynamic weighting factors}%
Table~\ref{table:ablation-DroneRGBT-12} and   \ref{table:ablation-rgbtcc-12}
show the effectiveness of dynamic weighting factors. In the table, ``w/o DW" refers to variants without dynamic weighting factors. Comparing the results of FusionCounting without dynamic weighting factors (w/o DW) to the original setting (Ours), it can be observed that the proposed dynamic weighting factors significantly improve both VIF and crowd counting performance.
Figure \ref{fig:loss} shows the comparisons of the training loss curves with and without dynamic weighting factors. It can be observed that the loss decreases significantly faster when using dynamic weighting factors compared to the case without them during the initial stage of training. This improvement can be attributed to the effective design of the weighting strategy, which balances the dependence between tasks. This observation also indicates that the introduction of dynamic weighting factors can more effectively guide the model to rapidly converge to the optimal solution.

Additionally, in Table \ref{Ablation_Z}, ablation experiments with parameter Z are shown on the model IADM, with values 1 to 3, and the best index is achieved when the parameter is 3.
\begin{table}[h]
\scriptsize
\centering
\caption{Ablation experiments on Z on the IADM model.}
\resizebox{0.48\textwidth}{!}{
\begin{tabular}{lccccc}
\hline
Z Value & GAME(0) & GAME(1) & GAME(2) & GAME(3) & RMSE  \\ \hline
1.0     & 22.69   & 30.25   & 36.62   & 48.22   & 37.84 \\
1.5     & 23.23   & 31.50   & 37.56   & 49.70   & 38.44 \\
2.0     & 22.06   & 30.27   & 37.16   & 48.94   & 35.75 \\
2.5     & 24.32   & 32.51   & 38.06   & 48.34   & 41.05  \\
3.0     & \cellcolor{lightred}\textbf{20.49}   & \cellcolor{lightred}\textbf{29.27}   & \cellcolor{lightred}\textbf{35.77}   & \cellcolor{lightred}\textbf{47.95}   & \cellcolor{lightred}\textbf{30.22} \\ \hline
\end{tabular}
}
\label{Ablation_Z}

\end{table}

\begin{figure}[!t]
	\centering
	\includegraphics[width=1\columnwidth]{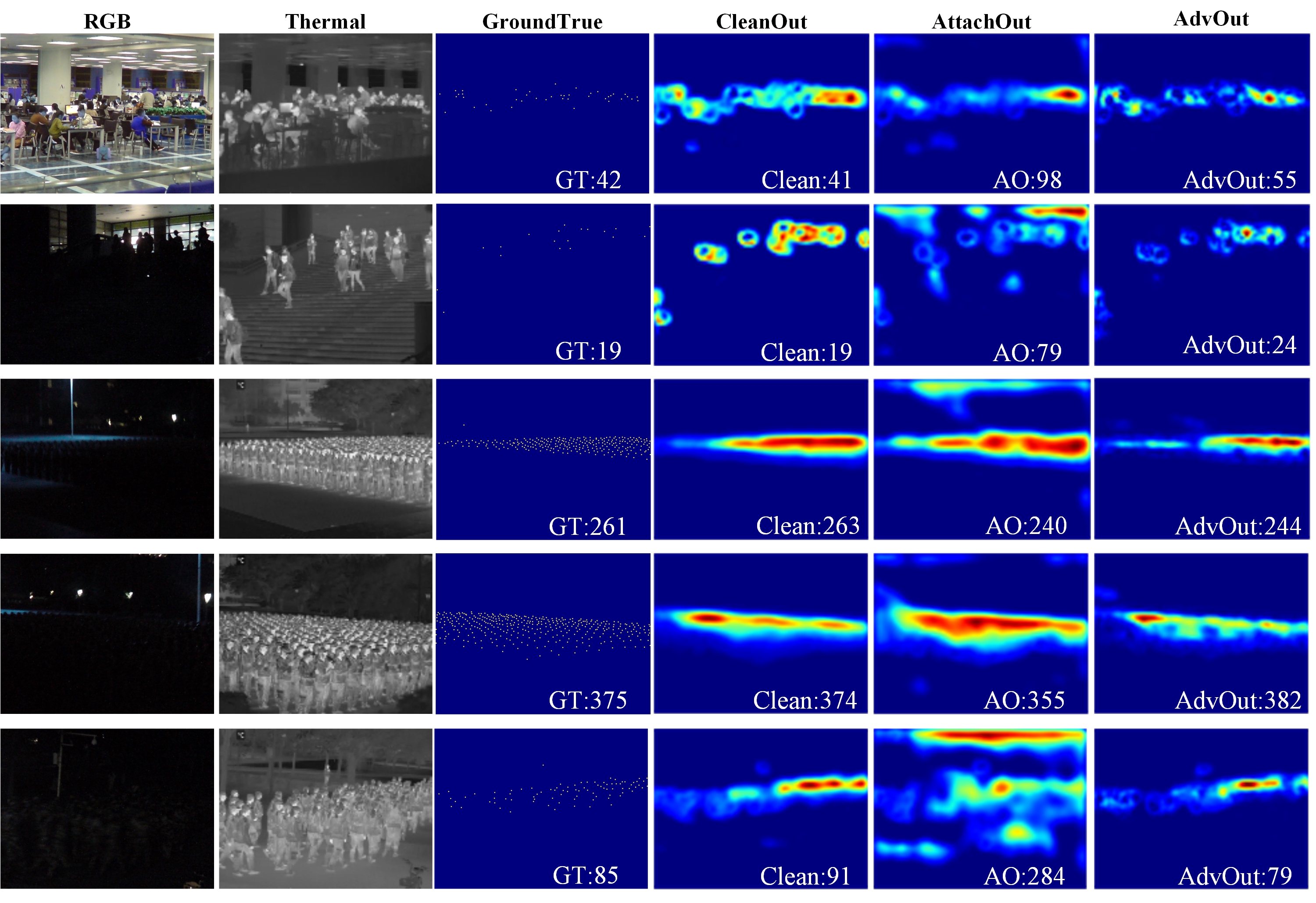}
	\caption{Crowd count density map visualization. The first and second columns show the RGB images and T images of different light intensity and crowd density scenes, the third column is GroundTrue, the fourth column is the density output of clean images (CleanOut), the fifth column is the density image after attack (AttackOut), and the last column shows the density image after adversarial training (AdvOut).}
	\label{fig:attack-counting}
\end{figure}

\begin{figure}[!t]
	\centering
	\includegraphics[width=1\columnwidth]{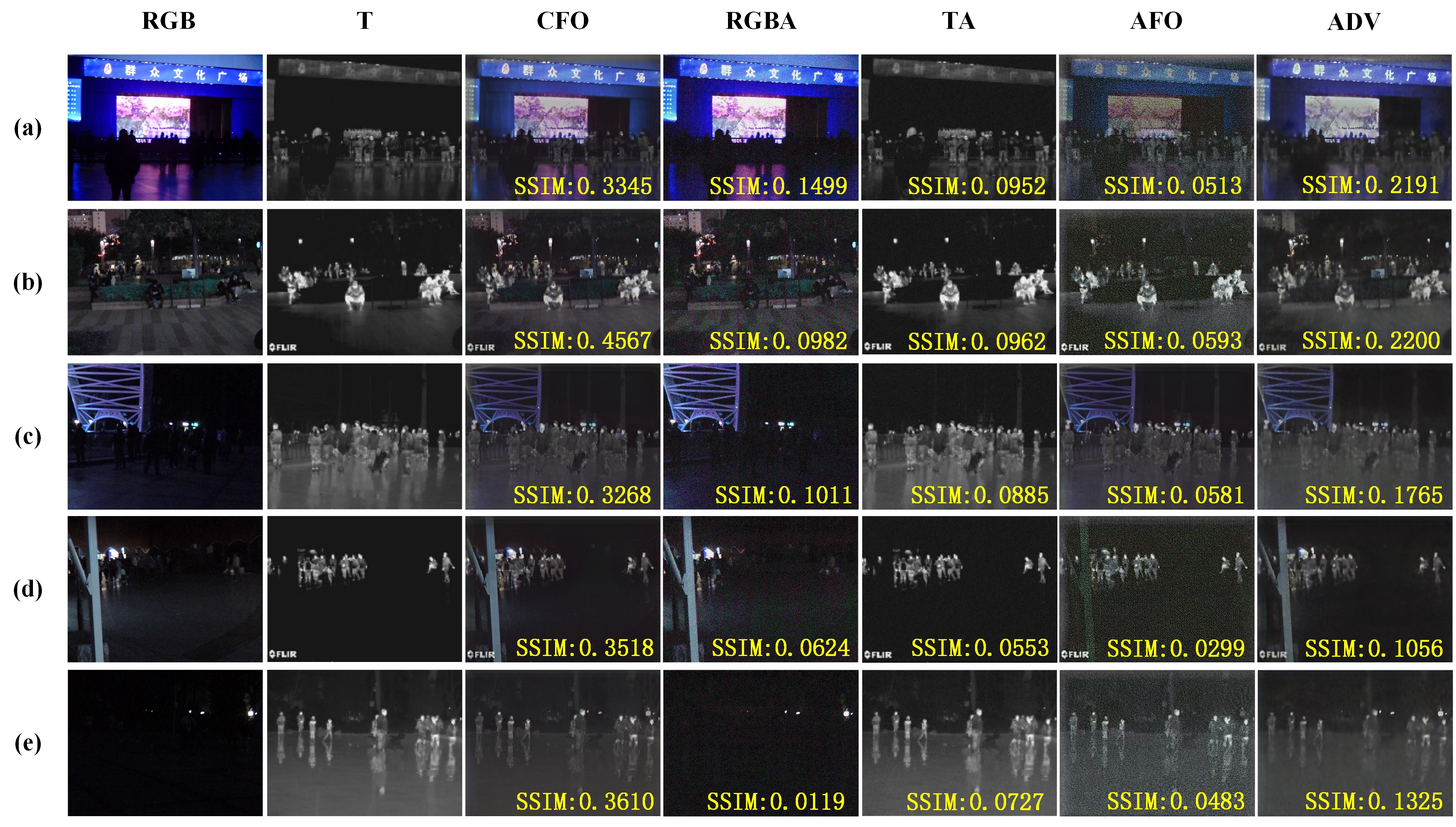}
	\caption{Analysis of the Attack and Defense Effects in Fusion Tasks. The third column presents the Clean Fusion Output (CFO) without any attack, while the fourth and fifth columns illustrate the outcomes of attacks on RGB images and T images, respectively. The sixth column displays the fusion output following an attack, and the seventh column exhibits the fusion output after adversarial training. These results demonstrate that adversarial training substantially enhances the robustness of the model.}
	\label{fig:attack-fusion}
\end{figure}

 \subsubsection{Effectiveness of adversarial training}%
\mbox{Table \ref{table:attack-ablation}} presents the results of ablation study on adversarial training.~Here, the input is attacked images.~It can be seen that introducing the adversarial learning makes our FusionCounting 
more robust to the adversarial attacks no matter what is the baseline.~Moreover, both the VIF and crowd counting performance is more robust after adversarial training. 

\textbf{Robustness analysis}. As shown in Fig.~\ref{fig:attack-counting} and Fig.~\ref{fig:attack-fusion}, adversarial attacks disrupt the density maps and significantly reduce the SSIM of visible and infrared images, leading to degraded fusion quality with increased noise and blur. In contrast, adversarially trained models produce density maps closer to the ground truth and improve fusion SSIM, resulting in clearer, less noisy images. These results demonstrate the effectiveness of adversarial training in enhancing the robustness and anti-interference capability of the model.
\begin{table*}[!h]
\centering

\caption{Attack effects and adversarial training for RGBT-CC. Our input is the attacked RGBT image, and we conduct attack experiment (attack) and adversarial experiment (Advtrain) on each baseline.~The input in these experiments is attacked images.}
\resizebox{1\textwidth}{!}{
\begin{tabular}{lc|cccccccc|ccccc}
\hline
\multirow{2}{*}{Baselines} & \multirow{2}{*}{Setting} & \multicolumn{8}{c|}{Visible-infrared image fusion} & \multicolumn{5}{c}{Crowd counting} \\ \cline{3-15}
& & Qabf$\uparrow$ & SSIM$\uparrow$ & PSNR$\uparrow$ & CC$\uparrow$ & SCD$\uparrow$ & SD$\uparrow$ & AG$\uparrow$ & SF$\uparrow$ & GAME(0)$\downarrow$ & GAME(1)$\downarrow$ & GAME(2)$\downarrow$ & GAME(3)$\downarrow$ & RMSE$\downarrow$ \\ \hline
\multirow{2}{*}{IADM} & w/o Advtrain & 0.27 & 0.49 & 60.72 & 0.25 & 0.31 & \cellcolor{lightred}\textbf{42.72} & 4.68 & 11.22 & 38.63 & 52.50 & 62.28 & 75.99 & 52.92 \\
 & Ours & \cellcolor{lightred}\textbf{0.30} & \cellcolor{lightred}\textbf{0.59} & \cellcolor{lightred}\textbf{61.63} & \cellcolor{lightred}\textbf{0.34} &\cellcolor{lightred}\textbf{0.45} & 36.56 & \cellcolor{lightred}\textbf{4.97} & \cellcolor{lightred}\textbf{12.00} & \cellcolor{lightred}\textbf{21.31} & \cellcolor{lightred}\textbf{33.07} & \cellcolor{lightred}\textbf{40.15} & \cellcolor{lightred}\textbf{53.18} & \cellcolor{lightred}\textbf{36.49} \\ \Xhline{1pt} 
 
\multirow{2}{*}{CSCA} & w/o Advtrain & 0.16 & 0.29 & 14.43 & 0.42 & 0.76 & 27.96 & \cellcolor{lightred}\textbf{8.04} & \cellcolor{lightred}\textbf{15.71} & 57.78 & 79.69 & 107.71 & 124.46 & 74.92 \\
 & Ours & \cellcolor{lightred}\textbf{0.30} & \cellcolor{lightred}\textbf{0.62} & \cellcolor{lightred}\textbf{15.59} & \cellcolor{lightred}\textbf{0.56} & \cellcolor{lightred}\textbf{1.54} & \cellcolor{lightred}\textbf{33.19} & 3.01 & 8.81 & \cellcolor{lightred}\textbf{32.95} & \cellcolor{lightred}\textbf{40.70} & \cellcolor{lightred}\textbf{48.22} &\cellcolor{lightred}\textbf{ 59.45} & \cellcolor{lightred}\textbf{52.32} \\ \Xhline{1pt} 
 
\multirow{2}{*}{MC$^3$Net} & w/o Advtrain & 0.29 & 0.54 & 15.55 & 0.49 & 1.24 & \cellcolor{lightred}\textbf{35.20} & \cellcolor{lightred}\textbf{5.86} & \cellcolor{lightred}\textbf{13.45} & 70.65 & 81.32 & 93.27 & 105.87 & 116.11 \\
 & Ours & \cellcolor{lightred}\textbf{0.30} & \cellcolor{lightred}\textbf{0.66} & \cellcolor{lightred}\textbf{15.67 }& \cellcolor{lightred}\textbf{0.51} & \cellcolor{lightred}\textbf{1.29} & 32.64 & 3.09 & 8.31 & \cellcolor{lightred}\textbf{39.08 }& \cellcolor{lightred}\textbf{49.14} & \cellcolor{lightred}\textbf{58.96} & \cellcolor{lightred}\textbf{70.91} & \cellcolor{lightred}\textbf{66.57} \\ \Xhline{1pt} 
 
\multirow{2}{*}{DEFNet} & w/o Advtrain & 0.17 & 0.15 & 12.84 & 0.36 & 0.87 & \cellcolor{lightred}\textbf{37.00 }& \cellcolor{lightred}\textbf{16.73} & \cellcolor{lightred}\textbf{41.09} & 316.06 & 325.53 & 340.18 & 371.96 & 410.32 \\
 & Ours & \cellcolor{lightred}\textbf{0.32} & \cellcolor{lightred}\textbf{0.63} & \cellcolor{lightred}\textbf{15.84} & \cellcolor{lightred}\textbf{0.58} & \cellcolor{lightred}\textbf{1.52} & 30.58 & 2.74 & 8.37 & \cellcolor{lightred}\textbf{67.73} & \cellcolor{lightred}\textbf{71.31} & \cellcolor{lightred}\textbf{75.67} & \cellcolor{lightred}\textbf{82.55} & \cellcolor{lightred}\textbf{104.93} \\ \hline
\end{tabular}
}
\label{table:attack-ablation}
\end{table*}

 \subsubsection{Effectiveness of the parallel architecture}
To verify the effectiveness of our parallel architecture, we also train a model using a serial structure for the VIF and crowd counting tasks—where the fusion decoder and counting decoder are cascaded. The comparison results are presented in Table~\ref{tab:compareAppVIF}. As shown, the parallel structure of FusionCounting significantly benefits the downstream crowd counting task. This suggests that learning both tasks concurrently allows better feature sharing and reduces error propagation between stages.

\begin{table}[!h]

\caption{Ablation studies on the architecture. `Series' denotes the series structure of two tasks in FusionCounting.}
\resizebox{1\columnwidth}{!}{
\begin{tabular}{lc|ccccc}
\hline
Baselines&Structure & GAME(0)$\downarrow$  & GAME(1)$\downarrow$  & GAME(2)$\downarrow$  & GAME(3)$\downarrow$  & RMSE$\downarrow$ \\ \hline
\multirow{2}{*}{IADM}&Series & 130.33 & 150.96 & 185.96 & 207.96 & 148.03 \\ 
&\textbf{Ours} & \cellcolor{lightred}\textbf{20.49}  & \cellcolor{lightred}\textbf{29.27}  & \cellcolor{lightred}\textbf{35.77}  & \cellcolor{lightred}\textbf{47.95}  & \cellcolor{lightred}\textbf{30.22}  \\ \Xhline{1pt} 
\multirow{2}{*}{CSCA}&Series&130.60&130.98&131.58&131.92&183.17\\
&\textbf{Ours}& \cellcolor{lightred}\textbf{21.23} & \cellcolor{lightred}\textbf{29.20} & \cellcolor{lightred}\textbf{36.00} & \cellcolor{lightred}\textbf{47.34} & \cellcolor{lightred}\textbf{35.23}\\ \Xhline{1pt} 
\multirow{2}{*}{MC$^3$Net}&Series&99.04&118.61&151.67&166.32&150.84\\
&\textbf{Ours}& \cellcolor{lightred}\textbf{15.84} & \cellcolor{lightred}\textbf{23.96} & \cellcolor{lightred}\textbf{32.50} & \cellcolor{lightred}\textbf{45.56} & \cellcolor{lightred}\textbf{26.02}\\ \Xhline{1pt} 
\multirow{2}{*}{DEFNet}&Series&102.25&116.94&139.49&150.27&162.70\\
&\textbf{Ours}& \cellcolor{lightred}\textbf{21.50} & \cellcolor{lightred}\textbf{36.46} & \cellcolor{lightred}\textbf{42.64} & \cellcolor{lightred}\textbf{53.47} & \cellcolor{lightred}\textbf{36.14}\\ \hline
\end{tabular}
}
\label{tab:compareAppVIF}
\end{table}

\subsection{Discussions}
\label{Discussions}
Tables \ref{table:ablation-rgbtcc-12} and \ref{table:ablation-DroneRGBT-12} show that the multi-task learning framework not only enhances image fusion quality but also improves counting performance through shared feature representation. As acknowledged, the counting head participates in training, sharing the backbone encoder with the fusion task—a standard practice in multi-task learning where auxiliary tasks provide additional supervision to guide feature learning and improve generalization. Specifically, within this shared encoder, features extracted at different levels capture local details, structural patterns, and global semantics. Low-level features preserve image edges and suppress noise, while high-level features provide abstract semantic information crucial for the counting task. The feature fusion module then gradually integrates these multi-level features, forming a richer and more comprehensive representation. This hierarchical fusion mechanism enhances information sharing and interaction between the tasks, allowing them to mutually reinforce each other. Importantly, while the counting head introduces valuable inductive bias during training, it does not directly contribute features or outputs to the fusion head, nor is it utilized during inference for the fusion task. Therefore, this design adheres to the multi-task learning paradigm, providing justified auxiliary supervision without introducing unfair advantage or information leakage. 

In summary, multi-task learning enables the fusion and counting tasks to complement each other effectively, providing a robust joint learning paradigm where the shared representation benefits both tasks.

\section{Conclusions}
In this study, we propose \textbf{FusionCounting}, a multi-task framework that connects two tasks in parallel through a shared encoder, enabling mutual reinforcement and optimizing both visible-infrared image fusion and crowd counting performance. To further improve performance, 
we design a dynamic loss weighting strategy that adaptively adjusts the weight ratio of current losses, accelerating network convergence and improving performance. Additionally, to enhance the robustness of the network, we incorporate an adversarial training strategy. By learning from attack data, the network mitigates its negative impact, strengthening the overall robustness of the multi-task framework. In future work, we will focus on addressing the misalignment between visible and infrared images, which remains a key challenge in cross-modal crowd counting. Our next step is to develop advanced alignment modules or explore alignment-robust representations, aiming to further improve the performance and practicality of FusionCounting.

\bibliography{fusion_counting}

\begin{thebibliography}{10}
\providecommand{\url}[1]{#1}
\csname url@samestyle\endcsname
\providecommand{\newblock}{\relax}
\providecommand{\bibinfo}[2]{#2}
\providecommand{\BIBentrySTDinterwordspacing}{\spaceskip=0pt\relax}
\providecommand{\BIBentryALTinterwordstretchfactor}{4}
\providecommand{\BIBentryALTinterwordspacing}{\spaceskip=\fontdimen2\font plus
\BIBentryALTinterwordstretchfactor\fontdimen3\font minus
  \fontdimen4\font\relax}
\providecommand{\BIBforeignlanguage}[2]{{%
\expandafter\ifx\csname l@#1\endcsname\relax
\typeout{** WARNING: IEEEtran.bst: No hyphenation pattern has been}%
\typeout{** loaded for the language `#1'. Using the pattern for}%
\typeout{** the default language instead.}%
\else
\language=\csname l@#1\endcsname
\fi
#2}}
\providecommand{\BIBdecl}{\relax}
\BIBdecl

\bibitem{10856398}
J.~Lei, J.~Li, J.~Liu \emph{et~al.}, ``{MLFuse: Multi-Scenario Feature Joint
  Learning for Multi-Modality Image Fusion},'' \emph{IEEE Transactions on
  Multimedia}, vol.~27, pp. 3880--3894, 2025.

\bibitem{10814643}
Q.~Xiao, H.~Jin, H.~Su \emph{et~al.}, ``{SPDFusion:A Semantic Prior
  Knowledge-Driven Method for Infrared and Visible Image Fusion},'' \emph{IEEE
  Transactions on Multimedia}, vol.~27, pp. 1691--1705, 2025.

\bibitem{zhang2023visible}
X.~Zhang and Y.~Demiris, ``Visible and infrared image fusion using deep
  learning,'' \emph{IEEE Transactions on Pattern Analysis and Machine
  Intelligence}, vol.~45, no.~8, pp. 10\,535--10\,554, 2023.

\bibitem{chen2024hitfusion}
J.~Chen, J.~Ding, and J.~Ma, ``Hitfusion: Infrared and visible image fusion for
  high-level vision tasks using transformer,'' \emph{IEEE Transactions on
  Multimedia}, vol.~26, pp. 10\,145--10\,159, 2024.

\bibitem{tang2022image}
L.~Tang, J.~Yuan, and J.~Ma, ``Image fusion in the loop of high-level vision
  tasks: A semantic-aware real-time infrared and visible image fusion
  network,'' \emph{Information Fusion}, vol.~82, pp. 28--42, 2022.

\bibitem{zhang2024mrfs}
H.~Zhang, X.~Zuo, J.~Jiang, C.~Guo, and J.~Ma, ``{MRFS: Mutually reinforcing
  image fusion and segmentation},'' in \emph{Proceedings of the IEEE/CVF
  Conference on Computer Vision and Pattern Recognition}, 2024, pp.
  26\,974--26\,983.

\bibitem{liu2022target}
J.~Liu, X.~Fan, Z.~Huang, G.~Wu, R.~Liu, W.~Zhong, and Z.~Luo, ``Target-aware
  dual adversarial learning and a multi-scenario multi-modality benchmark to
  fuse infrared and visible for object detection,'' in \emph{Proceedings of the
  IEEE/CVF Conference on Computer Vision and Pattern Recognition}, 2022, pp.
  5802--5811.

\bibitem{peng2022mfdetection}
Y.~Peng, G.~Liu, X.~Xu \emph{et~al.}, ``{MFDetection: A highly generalized
  object detection network unified with multilevel heterogeneous image
  fusion},'' \emph{Optik}, vol. 266, p. 169599, 2022.

\bibitem{liu2021cross}
L.~Liu, J.~Chen, H.~Wu, G.~Li, C.~Li, and L.~Lin, ``Cross-modal collaborative
  representation learning and a large-scale {RGBT} benchmark for crowd
  counting,'' in \emph{IEEE Conference on Computer Vision and Pattern
  Recognition}, 2021, pp. 4821--4831.

\bibitem{zhang2022spatio}
Y.~Zhang, S.~Choi, and S.~Hong, ``Spatio-channel attention blocks for
  cross-modal crowd counting,'' in \emph{Asian Conference on Computer Vision},
  2022, pp. 22--40.

\bibitem{zhou2024mc3}
W.~Zhou, X.~Yang, J.~Lei, W.~Yan, and L.~Yu, ``{MC$^3$Net}: Multimodality
  cross-guided compensation coordination network for {RGB-T} crowd counting,''
  \emph{IEEE Transactions on Intelligent Transportation Systems}, vol.~25,
  no.~5, pp. 4156--4165, 2024.

\bibitem{yang2024cagnet}
X.~Yang, W.~Zhou, W.~Yan, and X.~Qian, ``{CAGNet: Coordinated attention
  guidance network for RGB-T crowd counting},'' \emph{Expert Systems with
  Applications}, vol. 243, p. 122753, 2024.

\bibitem{ma2022swinfusion}
J.~Ma, L.~Tang, F.~Fan, J.~Huang, X.~Mei, and Y.~Ma, ``{SwinFusion:
  Cross-domain long-range learning for general image fusion via swin
  transformer},'' \emph{IEEE/CAA Journal of Automatica Sinica}, vol.~9, no.~7,
  pp. 1200--1217, 2022.

\bibitem{waghela2024robust}
H.~Waghela, J.~Sen, and S.~Rakshit, ``{Robust image classification: Defensive
  strategies against FGSM and PGD adversarial attacks},'' \emph{arXiv preprint
  arXiv:2408.13274}, 2024.

\bibitem{sener2018multi}
O.~Sener and V.~Koltun, ``Multi-task learning as multi-objective
  optimization,'' \emph{Advances in neural information processing systems},
  vol.~31, 2018.

\bibitem{bai2024ibfusion}
Y.~Bai, M.~Gao, S.~Li \emph{et~al.}, ``{IBFusion: An Infrared and Visible Image
  Fusion Method Based on Infrared Target Mask and Bimodal Feature Extraction
  Strategy},'' \emph{IEEE Transactions on Multimedia}, vol.~26, pp.
  10\,610--10\,622, 2024.

\bibitem{liu2024infrared}
J.~Liu, G.~Wu, Z.~Liu \emph{et~al.}, ``Infrared and visible image fusion: From
  data compatibility to task adaption,'' \emph{IEEE Transactions on Pattern
  Analysis and Machine Intelligence}, 2024.

\bibitem{zhao2023cddfuse}
Z.~Zhao, H.~Bai, J.~Zhang \emph{et~al.}, ``{CDDFuse: Correlation-driven
  dual-branch feature decomposition for multi-modality image fusion},'' in
  \emph{Proceedings of the IEEE/CVF conference on computer vision and pattern
  recognition}, 2023, pp. 5906--5916.

\bibitem{zhang2025texture}
K.~Zhang, L.~Sun, J.~Yan, W.~Wan, J.~Sun, S.~Yang, and H.~Zhang,
  ``Texture-content dual guided network for visible and infrared image
  fusion,'' \emph{IEEE Transactions on Multimedia}, vol.~27, pp. 2097 -- 2111,
  2025.

\bibitem{hu2024pfcfuse}
X.~Hu, Y.~Liu, and F.~Yang, ``{PFCFuse: A Poolformer and CNN fusion network for
  Infrared-Visible Image Fusion},'' \emph{IEEE Transactions on Instrumentation
  and Measurement}, 2024.

\bibitem{ma2019fusiongan}
J.~Ma, W.~Yu, P.~Liang, C.~Li, and J.~Jiang, ``{FusionGAN: A generative
  adversarial network for infrared and visible image fusion},''
  \emph{Information fusion}, vol.~48, pp. 11--26, 2019.

\bibitem{tang2024itfuse}
W.~Tang, F.~He, and Y.~Liu, ``{ITFuse: An interactive transformer for infrared
  and visible image fusion},'' \emph{Pattern Recognition}, vol. 156, p. 110822,
  2024.

\bibitem{liu2024stfnet}
Q.~Liu, J.~Pi, P.~Gao, and D.~Yuan, ``{STFNet: Self-supervised transformer for
  infrared and visible image fusion},'' \emph{IEEE Transactions on Emerging
  Topics in Computational Intelligence}, vol.~8, no.~2, pp. 1513--1526, 2024.

\bibitem{shi2025frefusion}
J.~Shi, P.~Duan, X.~Ma \emph{et~al.}, ``{Frefusion: Frequency Domain
  Transformer for Infrared and Visible Image Fusion},'' \emph{IEEE Transactions
  on Multimedia}, vol.~27, pp. 5722--5730, 2025.

\bibitem{yue2023dif}
J.~Yue, L.~Fang, S.~Xia, Y.~Deng, and J.~Ma, ``{Dif-Fusion: Toward high color
  fidelity in infrared and visible image fusion with diffusion models},''
  \emph{IEEE Transactions on Image Processing}, vol.~32, pp. 5705--5720, 2023.

\bibitem{yi2024diff}
X.~Yi, L.~Tang, H.~Zhang, H.~Xu, and J.~Ma, ``Diff-if: Multi-modality image
  fusion via diffusion model with fusion knowledge prior,'' \emph{Information
  Fusion}, vol. 110, p. 102450, 2024.

\bibitem{shopovska2019deep}
I.~Shopovska, L.~Jovanov, and W.~Philips, ``Deep visible and thermal image
  fusion for enhanced pedestrian visibility,'' \emph{Sensors}, vol.~19, no.~17,
  p. 3727, 2019.

\bibitem{zheng2024pedestrian}
B.~Zheng, H.~Huo, X.~Liu, S.~Pang, and J.~Li, ``Pedestrian detection-driven
  cascade network for infrared and visible image fusion,'' \emph{Signal
  Processing}, vol. 225, p. 109620, 2024.

\bibitem{liu2023multi}
J.~Liu, Z.~Liu, G.~Wu \emph{et~al.}, ``Multi-interactive feature learning and a
  full-time multi-modality benchmark for image fusion and segmentation,'' in
  \emph{Proceedings of the IEEE/CVF international conference on computer
  vision}, 2023, pp. 8115--8124.

\bibitem{fu2024segmentation}
H.~Fu, G.~Wu, Z.~Liu, T.~Yan, and J.~Liu, ``Segmentation-driven infrared and
  visible image fusion via transformer-enhanced architecture searching,'' in
  \emph{IEEE International Conference on Acoustics, Speech and Signal
  Processing}.\hskip 1em plus 0.5em minus 0.4em\relax IEEE, 2024, pp.
  4230--4234.

\bibitem{zhou2022defnet}
W.~Zhou, Y.~Pan, J.~Lei, L.~Ye, and L.~Yu, ``{DEFNet}: Dual-branch enhanced
  feature fusion network for {RGB-T} crowd counting,'' \emph{IEEE Transactions
  on Intelligent Transportation Systems}, vol.~23, no.~12, pp.
  24\,540--24\,549, 2022.

\bibitem{meng2024multi}
H.~Meng, X.~Hong, C.~Wang, M.~Shang, and W.~Zuo, ``Multi-modal crowd counting
  via a broker modality,'' in \emph{European Conference on Computer
  Vision}.\hskip 1em plus 0.5em minus 0.4em\relax Springer, 2024, pp. 231--250.

\bibitem{zhou2022domain}
K.~Zhou, Z.~Liu, Y.~Qiao, T.~Xiang, and C.~C. Loy, ``Domain generalization: A
  survey,'' \emph{IEEE Transactions on Pattern Analysis and Machine
  Intelligence}, vol.~45, no.~4, pp. 4396--4415, 2022.

\bibitem{gao2023alleviating}
C.~Gao, K.~Huang, J.~Chen \emph{et~al.}, ``Alleviating matthew effect of
  offline reinforcement learning in interactive recommendation,'' in
  \emph{Proceedings of the 46th international ACM SIGIR conference on research
  and development in information retrieval}, 2023, pp. 238--248.

\bibitem{hinton2015distilling}
G.~Hinton, ``Distilling the knowledge in a neural network,'' \emph{arXiv
  preprint arXiv:1503.02531}, 2015.

\bibitem{Peng2020Dronergbt}
T.~Peng, Q.~Li, and P.~Zhu, ``{RGB-T} crowd counting from drone: A benchmark
  and {MMCCN} network,'' in \emph{Proceedings of the Asian Conference on
  Computer Vision (ACCV)}, 2020, pp. 497--513.

\bibitem{jagalingam2015review}
P.~Jagalingam and A.~V. Hegde, ``A review of quality metrics for fused image,''
  \emph{Aquatic Procedia}, vol.~4, pp. 133--142, 2015.

\bibitem{xydeas2000objective}
C.~S. Xydeas and V.~Petrovic, ``Objective image fusion performance measure,''
  \emph{Electronics letters}, vol.~36, no.~4, pp. 308--309, 2000.

\bibitem{shah2013multifocus}
P.~Shah, S.~N. Merchant, and U.~B. Desai, ``Multifocus and multispectral image
  fusion based on pixel significance using multiresolution decomposition,''
  \emph{Signal, Image and Video Processing}, vol.~7, pp. 95--109, 2013.

\bibitem{aslantas2015new}
V.~Aslantas and E.~Bendes, ``A new image quality metric for image fusion: The
  sum of the correlations of differences,'' \emph{Aeu-international Journal of
  electronics and communications}, vol.~69, no.~12, pp. 1890--1896, 2015.

\bibitem{wang2004image}
Z.~Wang, A.~C. Bovik, H.~R. Sheikh, and E.~P. Simoncelli, ``Image quality
  assessment: from error visibility to structural similarity,'' \emph{IEEE
  Transactions on Image Processing}, vol.~13, no.~4, pp. 600--612, 2004.

\bibitem{cui2015detail}
G.~Cui, H.~Feng, Z.~Xu, Q.~Li, and Y.~Chen, ``Detail preserved fusion of
  visible and infrared images using regional saliency extraction and
  multi-scale image decomposition,'' \emph{Optics Communications}, vol. 341,
  pp. 199--209, 2015.

\bibitem{kong2024multiscale}
W.~Kong, H.~Li, and F.~Zhao, ``Multiscale modality-similar learning guided
  weakly supervised rgb-t crowd counting,'' \emph{IEEE Sensors Journal},
  vol.~24, no.~18, pp. 29\,121--29\,134, 2024.

\bibitem{Guerrero2015GAME}
R.-O. Guerrero-Gómez, B.~Torre-Jiménez, R.~López-Sastre,
  S.~Maldonado-Bascón, and D.~Oñoro-Rubio, ``Extremely overlapping vehicle
  counting,'' in \emph{Iberian Conference on Pattern Recognition and Image
  Analysis}, 2015, pp. 423--431.

\bibitem{li2021rfn}
H.~Li, X.-J. Wu, and J.~Kittler, ``{RFN-Nest: An end-to-end residual fusion
  network for infrared and visible images},'' \emph{Information Fusion},
  vol.~73, pp. 72--86, 2021.

\bibitem{wang2022res2fusion}
Z.~Wang, Y.~Wu, J.~Wang, J.~Xu, and W.~Shao, ``{Res2Fusion: Infrared and
  visible image fusion based on dense Res2net and double nonlocal attention
  models},'' \emph{IEEE Transactions on Instrumentation and Measurement},
  vol.~71, pp. 1--12, 2022.

\bibitem{wang2021unfusion}
Z.~Wang, J.~Wang, Y.~Wu, J.~Xu, and X.~Zhang, ``{UNFusion: A unified
  multi-scale densely connected network for infrared and visible image
  fusion},'' \emph{IEEE Transactions on Circuits and Systems for Video
  Technology}, vol.~32, no.~6, pp. 3360--3374, 2021.

\bibitem{qi2024ps}
J.~Qi, D.~E. Abera, and J.~Cheng, ``{PS-GAN: Pseudo Supervised Generative
  Adversarial Network With Single Scale Retinex Embedding for Infrared and
  Visible Image Fusion},'' \emph{IEEE Journal of Selected Topics in Applied
  Earth Observations and Remote Sensing}, 2024.

\bibitem{li2025fcdfusion}
H.~Li and Y.~Fu, ``{FCDFusion: A fast, low color deviation method for fusing
  visible and infrared image pairs},'' \emph{Computational Visual Media},
  vol.~11, no.~1, pp. 195--211, 2025.

\bibitem{tang2023rethinking}
L.~Tang, H.~Zhang, H.~Xu, and J.~Ma, ``Rethinking the necessity of image fusion
  in high-level vision tasks: A practical infrared and visible image fusion
  network based on progressive semantic injection and scene fidelity,''
  \emph{Information Fusion}, vol.~99, p. 101870, 2023.

\bibitem{cheng2024late}
J.~Cheng, C.~Feng, Y.~Xiao, and Z.~Cao, ``Late better than early: A
  decision-level information fusion approach for rgb-thermal crowd counting
  with illumination awareness,'' \emph{Neurocomputing}, vol. 594, p. 127888,
  2024.

\end{thebibliography}
\bibliographystyle{IEEEtran}

\vfill

\end{document}